\PassOptionsToPackage{notheorems}{jmlrutils}
\documentclass[12pt]{colt2026} %

\makeatletter
\let\@jmlr@check@packages\relax
\g@addto@macro\ps@jmlrtps{\let\@oddhead\@empty\let\@evenhead\@empty}
\makeatother

\usepackage{times}
\usepackage{amsmath,amssymb,amsthm,mathtools}
\usepackage{tikz}
\usetikzlibrary{fadings,patterns,shadows.blur,shapes,shapes.geometric,arrows.meta,positioning,fit,calc}
\usepackage{array,booktabs,multirow,tabularx}
\usepackage{yhmath,mathdots,extarrows}
\usepackage{siunitx,gensymb}
\usepackage{physics,cancel,color,xcolor,xspace}
\usepackage{mdframed,thmtools}
\usepackage{pifont}
\usepackage[normalem]{ulem} %
\usepackage{bbold}          %
\usepackage{enumitem}
\usepackage{wrapfig}
\usepackage{algorithm}      %
\usepackage{algpseudocode}  %

\usepackage{microtype}      %
\usepackage{needspace}      %
\usepackage{etoolbox}       %
\usepackage{caption}        %

\allowdisplaybreaks[2]

\preto\section{\needspace{4\baselineskip}}
\preto\subsection{\needspace{3\baselineskip}}
\preto\subsubsection{\needspace{3\baselineskip}}
\preto\paragraph{\needspace{2\baselineskip}}

\usepackage{hyperref}
\hypersetup{
    colorlinks=true,
    linkcolor=blue,
    citecolor=teal,
    urlcolor=magenta,
    pdfauthor={Konstantinos Emmanouilidis, Ziqing Xu, Salma Tarmoun, Lachlan MacDonald, Ren\'e Vidal},
    pdftitle={SGD with Large Learning Rates: General Analysis and Improved Rates},
    pdfborder={0 0 0}
}
\usepackage{url}
\usepackage{enumitem}
\usepackage[sort&compress,capitalize,nameinlink]{cleveref}		%

\definecolor{darkgreen}{RGB}{110,192,19}
\definecolor{shadecolor}{gray}{0.9}

\definecolor{thmblue}{RGB}{220,235,252}         %
\definecolor{thmblueborder}{RGB}{37,90,180}      %
\definecolor{lemmint}{RGB}{218,248,234}           %
\definecolor{lemmintborder}{RGB}{25,140,90}       %
\definecolor{propamber}{RGB}{255,244,215}         %
\definecolor{propamberborder}{RGB}{185,115,10}    %
\definecolor{assumlav}{RGB}{234,230,250}          %
\definecolor{assumlavborder}{RGB}{90,75,165}      %
\definecolor{examplebg}{RGB}{240,250,240}         %
\definecolor{exampleborder}{RGB}{60,160,60}       %

\newtheoremstyle{remark}
  {3pt}{3pt}{\itshape}{}{\bfseries}{.}{ }
  {Remark~\thmnumber{#2}}
\theoremstyle{remark}

\declaretheoremstyle[
  headfont=\normalfont\bfseries\color{thmblueborder},
  notefont=\normalfont\mdseries\color{thmblueborder},
  notebraces={(}{)},
  bodyfont=\normalfont,
  postheadspace=0.5em,
  spaceabove=6pt, spacebelow=4pt,
  mdframed={
    skipabove=6pt, skipbelow=6pt,
    linewidth=1.5pt,
    linecolor=thmblueborder,
    backgroundcolor=thmblue,
    innerleftmargin=8pt, innerrightmargin=8pt,
    innertopmargin=5pt, innerbottommargin=5pt,
    roundcorner=4pt}
]{thmbox}

\declaretheoremstyle[
  headfont=\normalfont\bfseries\color{lemmintborder},
  notefont=\normalfont\mdseries\color{lemmintborder},
  notebraces={(}{)},
  bodyfont=\normalfont,
  postheadspace=0.5em,
  spaceabove=6pt, spacebelow=4pt,
  mdframed={
    skipabove=6pt, skipbelow=6pt,
    linewidth=1.5pt,
    linecolor=lemmintborder,
    backgroundcolor=lemmint,
    innerleftmargin=8pt, innerrightmargin=8pt,
    innertopmargin=5pt, innerbottommargin=5pt,
    roundcorner=4pt}
]{lembox}

\declaretheoremstyle[
  headfont=\normalfont\bfseries\color{propamberborder},
  notefont=\normalfont\mdseries\color{propamberborder},
  notebraces={(}{)},
  bodyfont=\normalfont,
  postheadspace=0.5em,
  spaceabove=6pt, spacebelow=4pt,
  mdframed={
    skipabove=6pt, skipbelow=6pt,
    linewidth=1.5pt,
    linecolor=propamberborder,
    backgroundcolor=propamber,
    innerleftmargin=8pt, innerrightmargin=8pt,
    innertopmargin=5pt, innerbottommargin=5pt,
    roundcorner=4pt}
]{propbox}

\declaretheoremstyle[
  headfont=\normalfont\bfseries\color{assumlavborder},
  notefont=\normalfont\mdseries\color{assumlavborder},
  notebraces={(}{)},
  bodyfont=\normalfont,
  postheadspace=0.5em,
  spaceabove=6pt, spacebelow=4pt,
  mdframed={
    skipabove=6pt, skipbelow=6pt,
    linewidth=1.5pt,
    linecolor=assumlavborder,
    backgroundcolor=assumlav,
    innerleftmargin=8pt, innerrightmargin=8pt,
    innertopmargin=5pt, innerbottommargin=5pt,
    roundcorner=4pt}
]{assumbox}

\declaretheorem[style=assumbox,within=section]{definition}
\declaretheorem[style=thmbox,sibling=definition]{theorem}

\declaretheorem[style=lembox,sibling=definition]{assumption}
\declaretheorem[style=thmbox,sibling=definition]{corollary}
\declaretheorem[style=lembox,sibling=definition]{lemma}

\newcommand{\debug}[1]{#1}

\newcommand{\newmacro}[2]{\newcommand{#1}{\debug{#2}}}
\newcommand{\newop}[2]{\DeclareMathOperator{#1}{\debug{#2}}}

\newcommand{\R}{\mathbb{R}}

\newcommand{\E}{\operatorname{\mathbb{E}}}

\newop{\ex}{\mathbb{E}}     %
\newop{\prob}{\mathbb{P}}   %
\newop{\simplex}{\hull}

\newmacro{\noise}{U}
\newmacro{\filter}{\mathcal{F}}
\newmacro{\sgda}{\text{SGDA}}
\newmacro{\seg}{\text{SEG}}
\newcommand{\Expep}[1]{\ex\left[#1 \Big| \filter_t\right]}

\newmacro{\point}{x}
\newmacro{\pointalt}{\alt\point}
\newmacro{\pointaltalt}{\altalt\point}
\newmacro{\points}{\mathcal{X}}
\newmacro{\intpoints}{\relint\points}
\newmacro{\base}{p}
\newmacro{\basealt}{q}
\newmacro{\basealtalt}{u}
\newmacro{\open}{\mathcal{U}}
\newmacro{\closed}{\mathcal{C}}
\newmacro{\cpt}{\mathcal{K}}
\newmacro{\nhd}{\mathcal{U}}
\newmacro{\gmat}{g}
\newmacro{\gdist}{\dist_{\gmat}}
\newmacro{\mfld}{M}
\newmacro{\form}{\omega}
\newmacro{\tvec}{z}
\newmacro{\uvec}{u}
\newmacro{\basin}{\mathbb{B}}
\newmacro{\ball}{\basin}
\newmacro{\sphere}{\mathbb{S}}

\newmacro{\tstart}{0}		%
\newmacro{\timealt}{s}		%
\newmacro{\horizon}{T}		%

\newmacro{\traj}{x}		%
\newmacro{\trajalt}{y}		%
\newmacro{\trajaltalt}{z}		%

\newmacro{\flowmap}{\Theta}		%
\DeclarePairedDelimiterXPP{\flowof}[2]{\flowmap_{#1}}{(}{)}{}{#2}		%

\newop{\Opt}{Opt}		%
\newop{\Sol}{Sol}		%
\newop{\gap}{Gap}	
\newop{\dualitygap}{Duality-Gap}		%
\newop{\orcl}{Or}		%

\newmacro{\tfun}{f}		%
\newmacro{\obj}{f}		%
\newmacro{\objalt}{g}		%
\newmacro{\sobj}{F}		%

\newmacro{\gvec}{g}		%
\newmacro{\oper}{A}		%
\newmacro{\vecfield}{v}		%

\newcommand{\sol}[1][\point]{#1^{\ast}}		%
\newmacro{\solvec}{\vecfield(\sol)}		%
\newmacro{\solpay}{\eq[\payv]}		%

\newmacro{\signal}{V}		%
\newmacro{\step}{\gamma}		%
\newmacro{\learn}{\eta}		%

\newmacro{\vbound}{G}		%
\newmacro{\lips}{\ell}		%
\newmacro{\strong}{\mu}		%
\newmacro{\smooth}{\beta}		%

\newop{\tspace}{T}		%
\newop{\tcone}{TC}		%
\newop{\dcone}{\tcone^{\ast}}		%
\newop{\ncone}{NC}		%
\newop{\pcone}{PC}		%
\newop{\hull}{\Delta}		%

\newmacro{\cvx}{\mathcal{C}}		%
\newmacro{\subd}{\partial}		%

\newmacro{\minmax}{\mathcal{L}}		%

\newmacro{\minvar}{{\point_{1}}}		%
\newmacro{\minvaralt}{\alt\minvar}		%
\newmacro{\minvars}{\points_{1}}		%
\newmacro{\minsol}{\sol[\minvar]}		%

\newmacro{\maxvar}{\point_{2}}		%
\newmacro{\maxvaaltr}{\alt\maxvar}		%
\newmacro{\maxvars}{\points_{2}}		%
\newmacro{\maxsol}{\sol[\maxvar]}		%

\newop{\Eucl}{\Pi}		%
\newop{\logit}{\Lambda}		%
\newop{\dkl}{KL}		%

\newmacro{\hreg}{h}		%
\newmacro{\hconj}{\hreg^{\ast}}		%
\newmacro{\breg}{D}		%
\newmacro{\mprox}{P}		%
\newmacro{\mirror}{Q}		%
\newmacro{\fench}{F}		%
\newmacro{\depth}{H}		%
\newmacro{\hstr}{K}		%
\newmacro{\hker}{\theta}		%

\newmacro{\proxdom}{\points_{\hreg}}		%
\newmacro{\proxdomi}{\points_{\hreg_{\play}}}		%
\newmacro{\zone}{\mathbb{D}}		%

\DeclarePairedDelimiterXPP{\proxof}[2]{\mprox_{#1}}{(}{)}{}{#2}		%

\newmacro{\state}{x}
\newmacro{\statealt}{y}
\newmacro{\statealtalt}{z}
\newmacro{\solb}{\mathrm{R}}
\newmacro{\growth}{L}
\newmacro{\wqsmscale}{\mu}
\newmacro{\wqsmshift}{\lambda}

\newmacro{\stepalt}{\alpha}

\newmacro{\rvar}{g}
\newmacro{\kyrt}{\delta_{\text{\tiny{KYRT}}}}
\newmacro{\energy}{\mathcal{E}}

\newmacro{\irr}{\psi}

\newmacro{\tf}{\phi}

\newmacro{\normal}{\mathcal{N}}
\newmacro{\thres}{\theta}
\newmacro{\borel}{\mathcal{B}}

\DeclarePairedDelimiterX{\setdef}[2]{\lbrace}{\rbrace}{#1:#2}
\DeclarePairedDelimiterX{\inner}[2]{\langle}{\rangle}{#1,#2}
\DeclarePairedDelimiterXPP{\exclude}[1]{\mathopen{}\setminus}{\lbrace}{\rbrace}{}{#1}

\DeclarePairedDelimiterXPP{\exof}[1]{\ex}{\Big[}{\Big]}{}{%
   #1}
\DeclarePairedDelimiterXPP{\probof}[1]{\prob}{(}{)}{}{%
   #1}
\DeclarePairedDelimiterXPP{\oneof}[1]{\one}{\lbrace}{\rbrace}{}{%
   #1}

\newcommand{\Expe}[1]{\exof{#1}}

\DeclareMathOperator{\dist}{dist}

\DeclareMathOperator{\one}{\mathbb{1}} %

\DeclareMathOperator{\relint}{ri}

\usepackage{tcolorbox}
\newtcolorbox{yellowbox}{
  colback=yellow!7!white,
  colframe=blue!35!black,
  boxrule=0.4pt,arc=1.5pt,
  left=7pt,right=7pt,top=5pt,bottom=5pt,
  before skip=8pt,after skip=8pt,
  fontupper=\normalfont\normalsize
}

\title[SGD with Large Learning Rates]{SGD with Large Learning Rates: General Analysis \& Improved Rates}

\coltauthor{%
 \Name{Konstantinos Emmanouilidis} \Email{ekostas@engineering.upenn.edu}\\
 \addr University of Pennsylvania
 \AND
 \Name{Ziqing Xu} \Email{ziqingxu@wharton.upenn.edu}\\
 \addr University of Pennsylvania
 \AND
 \Name{Salma Tarmoun} \Email{starmoun@sas.upenn.edu}\\
 \addr University of Pennsylvania%
 \AND
 \Name{Lachlan MacDonald} \Email{lmacdo@seas.upenn.edu}\\
 \addr University of Pennsylvania
 \AND
 \Name{Ren{\'e} Vidal} \Email{vidalr@engineering.upenn.edu}\\
 \addr University of Pennsylvania
}

\begin{document}

\maketitle

\begin{abstract}%
Stochastic Gradient Descent lies at the core of modern machine learning. However its classical analysis typically restricts the stepsize to be below a curvature-dependent stability threshold. Recent developments in deterministic optimization have challenged this paradigm, showing that suitably chosen large stepsizes can not only preserve convergence, but also accelerate Gradient Descent (GD). Whether this phenomenon persists in the presence of stochastic gradient noise has remained an open question \citep{wu2024}.
In this work, we answer this question affirmatively for the cross-entropy loss with linear predictors and two-layer neural networks, establishing an accelerated \(\mathcal{O}(T^{-2})\) convergence rate. At the core of our analysis is a fine-grained characterization of the stochastic large-stepsize dynamics. This sharper control of the trajectory further yields tight convergence guarantees within the stable regime, improving existing bounds even in the deterministic setting. Taken together, our results show that the acceleration mechanism induced by large stepsizes survives stochasticity and extends beyond linear models. Numerical experiments demonstrate the benefits of large learning rates in practice.

\end{abstract}

\begin{keywords}%
  Edge of Stability, Large Stepsizes, SGD
\end{keywords}

\section{Introduction}

Stochastic Gradient Descent (SGD) and its variants constitute the workhorse of modern machine learning. From classical statistical learning problems to the training of large-scale neural networks, the basic recursion
$$
    w_{t+1}
    =
    w_t-\eta g_t,
    \qquad
    \E[g_t\mid w_t]
    =
    \nabla f(w_t),
$$
lies at the core of essentially every first-order stochastic optimization method. Despite this ubiquity, the current theoretical understanding of SGD is built around a remarkably conservative principle: the stepsize must be chosen sufficiently small so that each update is locally stable. Indeed, when \(f\) is \(L\)-smooth and the stochastic gradient has conditional variance bounded by \(\sigma^2>0\), the standard descent lemma gives
$$
    \E_t[f(w_{t+1})]
    \leq
    f(w_t)
    -
    \eta\left(1-\frac{L\eta}{2}\right)
    \|\nabla f(w_t)\|^2
    +
    \frac{L\eta^2}{2}\sigma^2.
$$
Thus, the canonical restriction \(\eta\lesssim 1/L\) emerges naturally: it guarantees that the dissipative contribution of the gradient dominates the curvature introduced by a single update. This simple principle, in one form or another, underlies much of the classical theory of stochastic approximation \citep{nesterov2018lectures}.

However, empirical work has shown that neural networks are routinely trained in the so-called \emph{Edge of Stability} (EoS) regime \citep{cohen2021gradient}, where the effective curvature approaches the stability threshold associated with the learning rate. In this regime, the loss does not decrease monotonically, the iterates exhibit persistent oscillations, and the classical descent picture no longer provides an adequate description of the dynamics. Despite that, training does not diverge. Quite the opposite: learning rates larger than those permissible by classical theory are frequently associated with markedly faster optimization than their conservative counterparts \citep{damian2023self, cohen2021gradient}.

These observations have prompted a sequence of theoretical works seeking to explain how optimization can remain effective beyond the classical stability threshold \citep{wu2023implicit, cai2024large}. In the logistic regression setting, this phenomenon is already visible in its simplest form. For linearly separable data, \citet{wu2024large} show that large-stepsize GD may traverse a non-monotone regime before eventually stabilizing, and despite this transient instability results in a provable acceleration of the optimization process. In particular, given a fixed step budget $T$, an appropriately chosen large constant stepsize improves the convergence rate from $\mathcal{O}\left(\frac{\log T}{T}\right)$ to $\mathcal{O}\left(\frac{\log T}{T^2}\right)$. Importantly, the acceleration is not achieved in spite of the unstable transient phase, but instead it is actually attributed to it.

This raises a natural question for stochastic optimization, since stochasticity changes the underlying dynamics in an essential way. More specifically, in the deterministic setting oscillations can be attributed directly to the interaction between the stepsize and the local curvature. In contrast, in SGD the trajectory is simultaneously affected by curvature-driven instability and gradient noise. Even when the learning rate lies within the classical stability range, individual stochastic updates need not decrease the objective, and even worse when the learning rate becomes large, separating stochastic fluctuations from genuine instability becomes considerably more delicate. As a result, the deterministic large-stepsize analysis does not transfer directly to the stochastic setting.

Recent works have taken a first step toward resolving this issue. \citet{wu2024large} provide upper bounds on logistic regression, characterizing the EoS regime. Simultaneously, \citet{emmanouilidis2026sgd} develop a stability framework for SGD with arbitrarily large stepsizes and characterize the dynamics by two qualitatively distinct regimes. Away from stability, the iterates exhibit an EoS phase characterized by non-monotone, curvature-driven behavior, while when the loss becomes sufficiently small, the dynamics are considered in the stable regime. Moreover, despite the persistent minibatch noise, the dynamics can return to the stable region and remain there with high probability. Hence, stochasticity does not preclude stabilization beyond the classical small-stepsize regime.

Stability, however, is only one aspect of the picture. The deterministic results reveal a considerably stronger phenomenon: the excursion beyond the classical stability threshold can lead to \emph{accelerated} convergence. Whether this acceleration mechanism survives the introduction of stochastic gradients remains unresolved and has been highlighted as a fundamental open question in the recent large-stepsize literature \citep{wu2024large, emmanouilidis2026sgd}. Standard SGD arguments offer little guidance in this regime because they are based precisely on the one-step descent inequality that large stepsizes invalidate. Conversely, deterministic EoS arguments rely on pathwise relations that are destroyed by the stochastic oracle. Establishing acceleration therefore requires a fundamentally different perspective: one that analyzes the stochastic trajectory globally, rather than attempting to enforce descent at each individual iteration.

The aforementioned discussion leads to the central question of this work:
\begin{center}
\textit{Can large stepsizes accelerate SGD?}
\end{center}

In this paper, we answer this question affirmatively for the cross-entropy loss setting. Our results show that the large-stepsize mechanism responsible for acceleration in deterministic cross-entropy optimization persists in the presence of stochastic gradients. More importantly, we identify the stochastic dynamics that make this possible: an initial large-learning-rate regime drives the process rapidly toward a region in which the same aggressive stepsize becomes stable, after which the trajectory admits substantially sharper convergence guarantees. In this sense, instability is not merely an artifact that SGD must overcome, it constitutes an essential ingredient of the route to acceleration.

\paragraph{Our contributions.}
Our contributions can be summarized as follows:
\begin{enumerate}%

\item \textbf{A sharp characterization of the large-stepsize dynamics.}
We provide a fine-grained analysis of SGD for the multi-class cross-entropy loss under arbitrary stepsizes \(\eta>0\). While previous work \citep{emmanouilidis2026sgd} allows for repeated transitions between stable and unstable phases, we show that the dynamics admit a substantially sharper description: once the iterates enter a sufficiently low-loss region, they remain in the stable regime with high probability. The key ingredient is a new supermartingale argument built around a suitably discounted loss threshold, which allows us to control the effect of stochastic fluctuations that could otherwise drive the trajectory out of stability. This gives the tight
$\mathcal{O}\left(\frac{1}{t}\right)$ convergence rate within the stable regime, matching the classical first-order rate for separable cross-entropy optimization \citep{nacson2019convergence}. %

\item \textbf{Acceleration of SGD through large stepsizes.}
Building on the aforementioned stability analysis, we establish for a fixed step budget $T$ an accelerated
$ \mathcal{O}\left(\frac{1}{T^2}\right)$ convergence rate for SGD with the cross-entropy loss on separable data. To the best of our knowledge, this is the first result showing that the large-stepsize acceleration mechanism survives in the presence of stochastic gradient noise. Moreover, in the full-batch limit, our bound removes the logarithmic factor appearing in prior deterministic guarantees and matches the canonical \(\mathcal{O}\left(\frac{1}{T^2}\right)\) accelerated rate associated with Nesterov's method \citep{Nesterov1983AMF}. Thus, acceleration can be obtained solely through an appropriate choice of the stepsize, without introducing momentum or modifying the underlying SGD iteration.

\item \textbf{Large-stepsize dynamics beyond linear predictors.}
We extend the theory to two-layer nearly homogeneous neural networks and provide a detailed description of their stochastic dynamics in the large-learning-rate regime. Despite the substantially more intricate geometry induced by the nonlinear parametrization, we recover the same qualitative phase structure as in the linear setting. In particular, we characterize the transition between the EoS and stable regimes and show that, once the loss becomes sufficiently small, the trajectory remains stable for all subsequent iterations with high probability. %

\item \textbf{Acceleration for two-layer neural networks.}
We further show that the acceleration mechanism is not an artifact of linear models. For two-layer nearly homogeneous neural-network predictors, we prove that an appropriate large-stepsize choice yields the accelerated rate $\mathcal{O}\left(\frac{1}{T^2}\right)$.
Hence, over a prescribed time horizon \(T>0\), the same rate as the celebrated accelerated gradient methods \citep{Nesterov1983AMF} can be attained by appropriately tuning the learning rate alone. No momentum term, extrapolation step, or modification of the SGD recursion is required. This demonstrates that stepsize-induced acceleration persists in a genuinely nonlinear stochastic setting.

\item \textbf{Empirical validation.}
Finally, we complement the theoretical analysis with extensive numerical experiments. The experiments corroborate the predicted transition from the EoS to the stable regime and consistently demonstrate the optimization benefits of large stepsizes across the settings covered by our theory.

\end{enumerate}

\section{Problem Setup \& Assumptions}

We consider the multi-class classification problem \eqref{eq:loss-def} with \(K\) classes, where \(x_i \in \mathbb{R}^d\) is the input, $y_i\in[K]$ denotes the corresponding label, $f$ is the predictor parameterized by weights \(W = [w_1,\ldots,w_K] \in \mathbb{R}^{d \times K}\), and the loss function is given by
\begin{equation}
    L(W)
    = \frac{1}{n}\sum_{i=1}^n \ell\big(f(x_i;W), y_i\big),
    \label{eq:loss-def}
    \vspace{-0.25cm}
\end{equation}
where $\ell(f,y)
    = \log\!\bigg(\sum_{j=1}^K e^{f_j}\bigg)
      - f_y$ is the classical cross-entropy loss function for the model $f$.
Our goal is to analyze the behavior of SGD in the large stepsize regime. For a mini-batch \(\mathcal{B}_t\) of size \(b \geq 1\){, consisting of $b$ indices sampled independently and uniformly from $[n]$ (with replacement)}, the SGD update at time \(t \geq 0\) is given by
\begin{equation}
    W_{t+1}
    = W_t - \eta\, g_t,
    \label{SGD} \tag{SGD}
\end{equation}
where $
    g_t
    = \frac{1}{b} \sum_{i\in\mathcal{B}_t}
    \nabla \ell(f(x_i;W_t), y_i)$ is the stochastic mini-batch gradient and $\eta>0$ is the stepsize of the algorithm. 

\textbf{Blanket Assumptions.} 
We, next, introduce the assumptions required for our theoretical results. 
We assume without loss of generality that inputs satisfy $\|x_i\|_2 \le 1, \forall i \in [n]$. This assumption can always be satisfied by simply normalizing the data. The next assumption concerns the separability of the dataset, which is common in the literature \citep{wu2023implicit, cai2024large} for this setting.

\begin{assumption}[Separable data]\label{assumpt: separable}
There exists \(W_* \in \mathbb{R}^{d\times K}\) with \(\|W_*\|_F = 1\) and a margin \(\gamma>0\) such that for every \(i\in[n]\) and every class \(j\neq y_i\), it holds that
$(W_*{^\top} x_i)_{y_i}
    \;-\;
    (W_*{^\top} x_i)_j
    \ge
    \gamma.$
\end{assumption}
Assumption~\ref{assumpt: separable} uses the multiclass margin of \citet{crammer2001algorithmic}: for the largest admissible $\gamma$, it equals the value of the hard-margin multiclass SVM,
\[
\max_{\|W\|_F\leq1}\min_{i\in[n]}\min_{j\neq y_i}\left[(W^\top x_i)_{y_i}-(W^\top x_i)_j\right].
\]

\section{Prior State of the Art}

We, first, review the existing results in the literature regarding the multi-class cross-entropy loss setting and then present our theoretical guarantees regarding acceleration. Recent work~\citep{emmanouilidis2026sgd} establishes that SGD alternates between an initial unstable regime, where a $\mathcal{O}\left(\frac{\ln{t}}{t}\right)$ rate for the best iterate for any stepsize $\eta > 0$ is shown, and a stable regime, which is attained when the loss is smaller than a certain threshold. More specifically, the stable set is defined as 
\begin{equation}
    \mathcal S
    :=
    \left\{
        W\in\mathbb{R}^{d\times K}:
        L(W)\leq\Tilde L
    \right\},
\end{equation}
where the stable threshold is given by $\Tilde{L}
    :=
    {\min\left\{
        \frac{1}{
            16\eta
            \left(
                1+\frac{2n}{b\gamma_0^2}
            \right)
        },
        \frac{1}{\sqrt2n\eta},
        \frac{1}{n}
    \right\}}$ with $\gamma_0:=\min\{\gamma,1\}$.
When the iterates are inside the stable set, the loss is shown to attain a last-iterate rate at a $\mathcal{O}\left(\frac{\ln{t}}{t}\right)$. However, the noise of the stochastic oracles may destabilize the dynamics, making the iterates exit momentarily the stable regime. For this reason, prior work \citep{emmanouilidis2026sgd} proves that the iterates will re-enter the stable set in a finite number of steps and will remain inside the stable regime with larger probability at every subsequent step as the loss steadily decreases.

\begin{theorem}[Stabilization of SGD, \citealt{emmanouilidis2026sgd}]\label{thm: self_stab}
    Let Assumption~\ref{assumpt: separable} hold. The iterates of~\eqref{SGD} with any stepsize $\eta > 0$ and batch size $b \geq 1$ satisfy the following.
    \begin{enumerate}[leftmargin=*]
        \item \textbf{Staying stable with high probability.}
        If $W_t \in \mathcal{S}$, then with probability at least $1-\delta'$ it holds that $W_{t+1} \in \mathcal{S}$, where $\delta'
        =
        \mathcal{O}\left(\exp\!\left(
        -\frac{b}{\eta L(W_t)^{3/2}}
        \right)\right).$

        \item \textbf{Return to stability.}
        Assume that the dynamics exit the stable regime at $t_{out}>0$, and fix $\delta\in(0,1)$. Then, with probability at least $1-\delta$, the iterates of~\eqref{SGD} return to the stable regime in at most $t_{\mathrm{re}} = \Tilde{\mathcal{O}}\left(\frac{\eta b^2}{\gamma^2 \delta \tilde{L}} \right)$ number of steps.
    \end{enumerate}
\end{theorem}

The aforementioned theorem is informative for the alternation between the stable and unstable regime. However, there is no guarantee that when the loss hits a certain threshold that the dynamics will remain \textit{forever} in the stable regime. This potential alternation between stable and unstable regimes is the main obstacle preventing prior analyses from proving an accelerated rate for SGD, motivating the fine-grained analysis of the next section.

\section{Convergence Guarantees}
Our results begin by providing a refined analysis of the phase transition phenomenon that occurs in the stochastic dynamics. More precisely, \citet{emmanouilidis2026sgd} show that the probability of remaining stable in the next iteration when the dynamics are already in the stable regime depends on the loss level and the distance of it from the stable level threshold $\Tilde{L}$. In this work, we prove the stronger statement that when the iterates attain a slightly lower loss threshold the dynamics remain stable for \emph{all} subsequent iterations. The intuition behind the aforementioned result is that since the noise in the cross-entropy setting is upper bounded by the loss, when the difference of the current loss from the stable threshold is sufficiently large, then with high probability the dynamics will not become unstable even in the presence of noise.

We, next, provide formally the result. 
For any $\delta\in(0,1)$, define the lower-level set $$\mathcal S_{\text{low}} = \left\{
        W\in\mathbb{R}^{d\times K}:L(W)\leq
        \frac{\delta}{3}\Tilde L
    \right\}$$ and the lower-level entrance time
$\tau_\delta
    :=
    \inf\left\{
        t\geq0:
        L(W_t)
        \leq
        \frac{\delta}{3}\Tilde L
    \right\}.$
\begin{theorem}
\label{thm:log_free_permanent_stability}
Let Assumption~\ref{assumpt: separable} hold.
Consider \ref{SGD} with stepsize $\eta>0$ and batch size $b\geq1$ and fix $\delta \in (0, 1)$. Then, the following hold with
probability at least $1-\delta$, 
\begin{enumerate}
    \item The iterates enter the lower-level set in $\tau_\delta
    =
    \Tilde{\mathcal{O}}\left(\frac{1+\eta}{\gamma_0^2 \eta\delta^2\Tilde L}\right)$ iterations.
    {\item For every iteration $t \geq \tau_\delta$, the iterates remain in the stable set $W_t\in\mathcal S$.
    \item For any fixed $T >\tau_\delta$, then}
    \begin{equation}
    L(W_T)
    \leq
    \frac{
        24
    }{
        \gamma_0^2
        \eta\delta(T-\tau_\delta)
    }.
    \label{eq:log_free_last_iterate_rate}
\end{equation}
\end{enumerate}
\end{theorem}

Theorem~\ref{thm:log_free_permanent_stability} provides a refined characterization of the dynamics at the stable regime. In particular, it establishes that after a certain number of iterations the dynamics will not only become stable, but they will enter with high probability the lower-level set. The aforementioned fact is of particular interest as the second point of the theorem provides a $\mathcal{O}\left(\frac{1}{t}\right)$ rate after entering the low-level set. We note that the established rate of convergence is strictly better than the $\mathcal{O}\left(\frac{ln t}{t}\right)$ rate of prior results \citep{emmanouilidis2026sgd} {and matches, for a fixed stepsize, the $\mathcal{O}\left(\frac{1}{t}\right)$ rate of small-stepsize gradient descent on separable data \citep{soudry2018implicit, nacson2019convergence}.} %

With the refined analysis at hand, we can now provide the acceleration of SGD with large stepsizes. 
\begin{corollary}
\label{cor:log_free_acceleration}
Let Assumption~\ref{assumpt: separable} hold and fix $\delta \in (0, 1)$. {There exist $c_0>0$ and $T_0(c)>0$ such that for any $c\in(0,c_0]$ and $T\geq T_0(c)$,} \ref{SGD} with stepsize {$\eta=cT$} and batch size $b\geq1$ satisfies with probability at least $1-\delta$ that
\begin{equation}
    L(W_T)
    \leq
    {\frac{
        48
    }{
        \delta\gamma^2_0 c T^2
    }}
    =
    \mathcal{O}\left(
        \frac1{T^2}
    \right).
\label{eq:log_free_acceleration}
\end{equation}
\end{corollary}

Corollary~\ref{cor:log_free_acceleration} establishes an $\mathcal{O}\left(\frac{1}{T^2}\right)$ accelerated rate with high probability for SGD. This improves upon the $\mathcal{O}\left(\frac{1}{T}\right)$ of classical SGD with small learning rates \citep{nacson2019stochastic} and is a characteristic of the large stepsize regime. In the full-batch setting, as a special case, our result improves the prior deterministic result of $\mathcal{O}\left(\frac{lnT}{T^2}\right)$ rate \citep{wu2024large}, shaving off unnecessary logarithmic factors and making the result match the rate of Nesterov acceleration \citep{Nesterov1983AMF}.

\section{Large Stepsizes in Two-Layer Neural Networks}

A natural question is whether the previously established accelerated rate of convergence persists beyond linear models and extends to the neural-network setting encountered in practice. This extension is far from immediate as the geometry of the loss landscape becomes substantially more intricate, and the curvature now depends nonlinearly on both the network parameters and the data. Nevertheless, many commonly used architectures retain sufficient structure, such as approximate homogeneity and controlled derivatives of the activation function, to allow a precise analysis of their stochastic dynamics. In this section, we show that, under mild regularity conditions on the activation function, the qualitative picture developed above remains the same for two-layer neural networks trained with the cross-entropy loss.

We consider two-layer neural networks in mean-field scaling with the predictor given by
\begin{equation}
   \label{eq: 2nn}
    f(W; x)
    :=
    \frac{1}{m}\sum_{j=1}^m a_j \phi(x^\top W^{(j)}),
\end{equation}
where the coefficients $a_j \in \{\pm 1\}$ are fixed and
$W = (W^{(j)})_{j=1}^m \in \R^{md\times K}$ are trainable parameters.
We run SGD with the rescaled stepsize $\tilde\eta := m \eta$, namely $W_{t+1} = W_t - \tilde\eta g_t$, to train the network.
Without loss of generality, we assume $\sup_i \|x_i\|\leq 1$ throughout the section. 

Following the prior literature \citep{emmanouilidis2026sgd}, we impose the following conditions on the activation function of the neural network.

\begin{assumption}
[Activation function conditions]
\label{assump: activation}
Let $\phi:\R \to \R$ be continuously differentiable and satisfy the following
\begin{enumerate}[leftmargin=*]
    \item \label{assump: activation:grad}
    \textbf{Derivative condition.}
    There exists $\alpha \in \left(\frac{1}{1+\gamma},1\right)$ such that
    $\alpha \leq \phi'(z) \leq 1$.
    \item \label{assump:activation:smooth}
    \textbf{Smoothness.}
    There exists $\tilde\beta>0$ such that, for all $x,y\in\R$,
    $|\phi'(x)-\phi'(y)| \leq \tilde\beta |x-y|$.

    \item \label{assump: activation:near-homogeneous}
    \textbf{Near-homogeneity.}
    There exists $\kappa>0$ such that, for every $z\in\R$,
    $|\phi(z)-\phi'(z)z| \leq \kappa$.
\end{enumerate}
\end{assumption}

Assumption~\ref{assump: activation} is satisfied by several common activations after a mild leaky modification. The smoothness and near-homogeneity conditions are classical in analyses of optimization for two-layer networks, and are also used in prior work on large-stepsize (stochastic) gradient descent~\citep{cai2024large, emmanouilidis2026sgd}. The derivative condition ensures that during training every neuron continues to receive a non-trivial gradient, ruling out the degenerate case when some neurons become frozen.
The lower bound $\alpha > 1/(1+\gamma)$ ensures that the effective margin $\tilde\gamma
  :=
  \alpha\gamma - (1-\alpha)$ remains positive. We note that $\tilde\gamma$ plays the role of the margin in the two-layer network analysis and can be interpreted as the original margin $\gamma$, discounted by the smallest possible activation derivative, and corrected by the variation of the activation derivatives across classes. In the limit $\alpha\to 1$, $\tilde\gamma\to\gamma$, recovering the usual margin in binary logistic regression.

The next lemma indicates that Assumption~\ref{assump: activation} is not restrictive, as leaky versions of standard activation functions satisfy the assumption.

\begin{lemma}[\citealt{emmanouilidis2026sgd}]
\label{lem:leaky-activations}
For a margin $\gamma > 0$, let $c$ satisfy $1/(1+\gamma)<c<1$.
The following leaky variants of common activation functions satisfy Assumption~\ref{assump: activation}.
\begin{itemize}[leftmargin=*]
\item \textbf{GELU, Softplus, SiLU.}
Let $\phi$ be GELU, Softplus, or SiLU, and define $\widetilde\phi(x)
    :=
    \frac{1+2c}{3}x+\frac{1-c}{6}\phi(x)$.
Then $\widetilde\phi$ satisfies
Assumption~\ref{assump: activation} with
$\alpha=c$, $\widetilde\beta=4(1-c)$, and $\kappa=1-c$.
\item \textbf{Huberized ReLU.}
Let $\phi$ be the Huberized $\mathrm{ReLU}_h$. Then,
$\tilde\phi(x) := c x + (1-c)\phi(x)$
satisfies Assumption~\ref{assump: activation} with
$\alpha=c$, $\tilde\beta=(1-c)/h$, and $\kappa=(1-c)h/2$.
\item \textbf{Tanh, Sigmoid.}
The leaky tanh
$\tilde\phi(x) := c x + (1-c)\tanh(x)$
and the leaky sigmoid
$\tilde\phi(x) := c x + \frac{1-c}{1+e^{-x}}$
both satisfy Assumption~\ref{assump: activation} with
$\alpha=c$, $\tilde\beta=1-c$, and $\kappa=1-c$.
\end{itemize}
\end{lemma}

With those assumptions at hand, we now state the main convergence results for two-layer neural networks. It is known \citep{emmanouilidis2026sgd} that in the EoS regime SGD attains a $\mathcal{O}\left(\frac{lnt}{t}\right)$ rate in terms of the best iterate of the loss. In the stable regime, defined as
\[
    \mathcal S^{NN}
    :=
    \left\{
        W\in\mathbb{R}^{md\times K}:
        L(W)\leq\Tilde L^{NN}
    \right\}.
\]
with the
stable threshold being
\begin{equation}
    \widetilde L^{NN}
    =
    \min\left\{
    \frac{1}{
    {16}\eta(1+\widetilde\beta)
    \left(
    1+\frac{2n}{
    b\min\{\widetilde\gamma^2,1\}}
    \right)},
    \frac{1}{\sqrt{2}n\eta},
    \frac{1}{2ne^{\kappa+2}}
    \right\}.
\label{eq:NN_quantitative_stable_threshold}
\end{equation}
and $\tilde\gamma_0:=\min\{\tilde\gamma,1\}$.

SGD with any stepsize $\eta > 0$ converges in last-iterate sense at a rate $\mathcal{O}\left(\frac{lnt}{t}\right)$. However, prior work leaves open the possibility that the dynamics are momentarily destabilized, becoming unstable for a certain number of iterations. Our results start by refining the transition between stable and unstable regimes, leading to improved convergence  guarantees in the stable regime.

Following the intuition from the cross-entropy setting with a linear model, we define for a fixed $\delta \in (0, 1)$ the lower-level set for the neural network dynamics
\[
    \mathcal S_{low}^{NN}
    :=
    \left\{
        W\in\mathbb{R}^{md\times K}:
        L(W)\leq\frac{\delta}{3}
        \Tilde L^{NN}
    \right\}.
\] and the corresponding entrance time
$\tau_\delta^{NN}
    :=
    \inf\left\{
        t\geq0:
        L(W_t)
        \leq
        \frac{\delta}{3}
        \Tilde L^{NN}
    \right\}.$

\begin{theorem}
\label{thm:NN_log_free_permanent_stability}
Let Assumptions~\ref{assumpt: separable}
and~\ref{assump: activation} hold. Consider \ref{SGD} 
for the two-layer network in~\eqref{eq: 2nn} and fix $\delta \in(0, 1)$.
Then, with probability at
least $1-\delta$, the following hold:
\begin{enumerate}
    \item The iterates enter the {lower-level set $\mathcal S_{low}^{NN}$} in at most $\tau_\delta^{NN}
    =
    \Tilde{\mathcal{O}}\left(\frac{1+\eta}{\tilde\gamma_0^2 \eta\delta^2\Tilde L^{NN}}\right)$ iterations.
    {\item For every $t \geq \tau_\delta^{NN}$ it holds that
$W_t\in\mathcal S^{NN}$.
    \item For any fixed $T>\tau_\delta^{NN}$, then the last iterate satisfies}
\begin{equation}
    L(W_T)
    \leq
    \frac{
        24
    }{
        \delta\tilde\gamma_0^2\eta
        (T-\tau_\delta^{NN})
    }.
\label{eq:NN_log_free_last_iterate}
\end{equation} 
\end{enumerate}
\end{theorem}

Theorem~\ref{thm:NN_log_free_permanent_stability} establishes that after a certain number of iterations with high probability the dynamics will have entered the lower-level set and thus they will be stable. More importantly, the second point of the theorem provides a last iterate convergence guarantee for the iterates of SGD with any positive stepsize $\eta > 0$. Ensuring that the dynamics will reach not only the stable set but also the lower-level set allows us to obtain a rate $\mathcal{O}\left(\frac{1}{t}\right)$, strictly outperforming the prior analyses \citep{emmanouilidis2026sgd}. As a special case, for full-batch GD the aforementioned guarantee is tighter than prior analysis of two-layer neural networks for binary classification \citep{cai2024large}. 

We highlight that our analysis through the lower-level set is the key in attaining an improved rate among the aforementioned prior works. More specifically, the use of the correct level set in combination with a Lyapunov stability analysis allows us to establish a tighter rate of convergence for the aforementioned setting.

We, next, establish that the use of large stepsizes leads to an accelerated rate of convergence in the case of two-layer neural networks.
\begin{corollary}
\label{cor:NN_log_free_acceleration}
Let Assumptions~\ref{assumpt: separable} and
\ref{assump: activation} hold and fix $\delta\in(0,1)$. Consider SGD for the two-layer NN of \eqref{eq: 2nn} with batch
size $b\geq1$ and scaled stepsize $\widetilde\eta=m\eta$, {where $\eta=cT$ with $c\in(0,c_0^{NN}]$ and $T\geq T_0^{NN}(c)$, for $c_0^{NN},T_0^{NN}(c)>0$.}
Then, with probability at least $1-\delta$, the last iterate satisfies
\begin{equation}
    L(W_T)
    \leq
    {\frac{48}{
        \delta\Tilde\gamma_0^2 c T^2
    }}
    =
    \mathcal O\left(
        \frac1{T^2}
    \right).
    \label{eq:NN_log_free_acceleration}
\end{equation}
\end{corollary}

Corollary~\ref{cor:NN_log_free_acceleration} establishes that SGD with large stepsizes converges at an accelerated rate of $\mathcal{O}\left(\frac{1}{T^2}\right)$. This is the nonlinear analogue of the acceleration guarantee obtained for the linear cross-entropy model. We emphasize that in the special of full-batch GD our result is tighter than the prior state-of-the-art $\mathcal{O}\left(\frac{lnT}{T^2}\right)$ rate \citep{cai2024large}.

\section{Experiments}
\label{sec:experiments}

In this section, we empirically validate the theoretical results established in the preceding sections for linear models and two-layer neural networks trained with SGD using large step sizes. First, the training dynamics initially enter the EOS regime before transitioning to a stable regime. Second, when the step size is chosen proportional to the training horizon $T$, the empirical mean final loss decays at the accelerated rate $\mathcal O(T^{-2})$, consistent with the predictions of \cref{cor:log_free_acceleration,cor:NN_log_free_acceleration}.
\begin{figure}[h]
    \centering
    \includegraphics[width=0.9\linewidth]{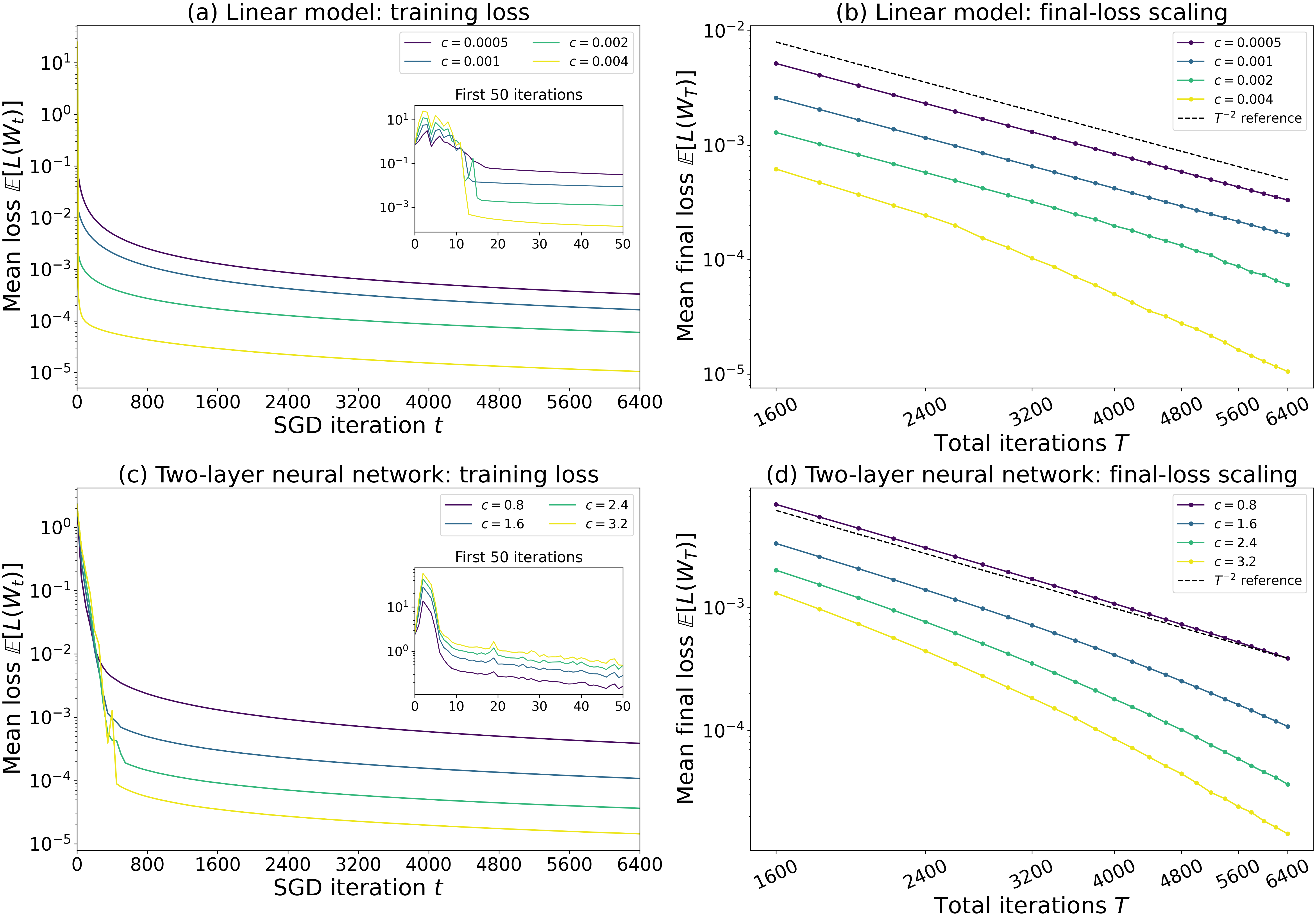}
    \caption{Training dynamics and final-loss scaling for the linear model (top) and two-layer model on frozen ConvNeXt features (bottom). Left panels show mean training loss at $T=6400$, with the first 50 iterations enlarged in the insets. Right panels show mean final loss versus $T$ under $\eta=cT$. Dashed lines indicate the $T^{-2}$ reference rate which is derived in ~\cref{cor:log_free_acceleration} and \cref{cor:NN_log_free_acceleration}. Averages are over independent minibatch-sampling streams with the data and features held fixed.}
    \label{fig:linear-2layer-nn}
\end{figure}
\paragraph{Setup.}
We consider a linear model on synthetic two-class dataset and a two-layer neural network on frozen CIFAR-10 features~\citep{krizhevsky2009learning}. For the linear model, we draw \(50\) examples per class from \(\mathcal{N}(\mu_k,0.025^2I_2)\), with \(\mu_1=(\pi/3,1)\) and \(\mu_2=(2\pi/3,1)\).
For CIFAR-10, we select a balanced subset of \(2{,}000\) training images, extract and normalize features using an ImageNet-pretrained ConvNeXt-Tiny~\citep{DBLP:journals/corr/abs-2201-03545}, and train a width-\(16\) fixed-output two-layer neural network with activation $\phi(z)\!=\!0.95z+0.05\tanh(z)$. Both models are initialized at zero and trained using cross-entropy loss and mini-batch SGD with batch size \(80\). For each \(T\in\{1600,1800,\ldots,6400\}\), we set the step size \(\eta=cT\), using \(c\in\{0.0005,0.001,0.002,0.004\}\) for the linear model and \(c\in\{0.8,1.6,2.4,3.2\}\) for the two-layer model. We estimate expectations over SGD sampling by averaging the full training loss over \(50\) and \(10\) independent mini-batch streams, respectively, while keeping the data and all fixed model components unchanged.

At the fixed horizon \(T=6400\), plots (a) and (c) in \cref{fig:linear-2layer-nn} show that large-step-size SGD initially exhibits sharp loss increases and non-monotone oscillations, characteristic of the EoS regime. After this transient phase, the loss enters a stable regime and decreases steadily. Plots (b) and (d) examine the acceleration obtained by choosing the step size proportional to the training horizon, \(\eta=cT\). For sufficiently large \(T\), the expected final-loss curves are approximately linear on the log-log scale and decay at a rate comparable to the \(T^{-2}\) reference, consistent with the rates established in \cref{cor:log_free_acceleration,cor:NN_log_free_acceleration}. 
Taken together, these results empirically verify the EoS-to-stability transition and the large-step-size acceleration predicted by our theory for both linear and nonlinear models.

\section{Conclusion}

In this work, we showed that large stepsizes can play a constructive role in stochastic optimization, establishing accelerated \(\mathcal{O}(T^{-2})\) convergence for SGD with the cross-entropy loss for both linear models and two-layer neural networks. Our analysis provides a refined characterization of the transition from the Edge-of-Stability to a stable low-loss regime and shows that, once this region is reached, the dynamics remain stable with high probability despite persistent stochastic noise. These results suggest that the classical stability threshold should not be viewed as an absolute barrier: in structured learning problems, crossing it can lead to faster optimization without momentum or any modification of the underlying SGD iteration. An important direction for future work is to understand how broadly this instability-to-acceleration mechanism extends to deeper architectures and more general stochastic learning problems.

\newpage
\bibliography{sections/ref}

\appendix

\newpage
\addcontentsline{toc}{section}{Appendix}
\tableofcontents
\newpage

\section{Proof Roadmap}

\begin{figure}[h!]
\centering
\begin{tikzpicture}[
    font=\small,
    >=Latex,
    arrow/.style={
        ->,
        semithick,
        draw=black!65
    },
    common/.style={
        draw=black!45,
        rounded corners=2pt,
        fill=gray!7,
        align=left,
        inner sep=6pt,
        text width=.90\linewidth
    },
    linear/.style={
        draw=blue!45!black,
        rounded corners=2pt,
        fill=blue!5,
        align=left,
        inner sep=6pt,
        text width=.39\linewidth
    },
    neural/.style={
        draw=teal!50!black,
        rounded corners=2pt,
        fill=teal!5,
        align=left,
        inner sep=6pt,
        text width=.39\linewidth
    },
    result/.style={
        draw=orange!60!black,
        rounded corners=2pt,
        fill=orange!7,
        align=left,
        inner sep=6pt,
        text width=.39\linewidth
    },
    node distance=4mm and 10mm
]

\node[common] (setup) {
\textbf{Common setup.}
Multiclass cross-entropy, separability
(Assumption~\ref{assumpt: separable}), with-replacement minibatch SGD,
and the potentials
$G(W)\leq L(W)\leq F(W)$.
};

\node[
    linear,
    below left=of setup.south,
    anchor=north east,
    xshift=-2mm
] (linprep) {
\textbf{Linear predictors: preparatory lemmas.}
Lemma~\ref{lemma: self_boundedness}: self-bounding and margin inequalities.\\
Lemma~\ref{lemma: average_G_bound}: average surrogate-loss bound.\\
Lemma~\ref{lemma: stochastic_grad_bound}: stochastic-gradient moments.
};

\node[
    neural,
    below right=of setup.south,
    anchor=north west,
    xshift=2mm
] (nnprep) {
\textbf{Two-layer networks: preparatory lemmas.}\\
Lemma~\ref{lem:nn-grad-bound}: gradient bounds.\\
Lemma~\ref{lem:nn-hessian}: curvature bounds. \\
Lemma~\ref{lem:nn-G-cesaro}: average surrogate-loss bound.
};

\node[linear, below=of linprep] (linentry) {
\textbf{Entrance and stable-regime drift.}
The average surrogate-loss bound and Markov's inequality give
$\tau_\delta<t_{\max}$.\\
The structural and moment bounds give the quadratic drift in
Lemma~\ref{lem:linear-stable-descent}.
};

\node[neural, below=of nnprep] (nnentry) {
\textbf{Entrance and stable-regime drift.}
The average surrogate-loss bound and Markov's inequality give
$\tau_\delta^{NN}<t_{\max}^{NN}$.\\
The structural estimates give the quadratic drift in Lemma~\ref{lem:nn-descent}.
};

\node[linear, below=of linentry] (linproc) {
\textbf{Stopped and killed processes.}
The stopped process and Doob's inequality yield permanent stability.\\
Lemma~\ref{lem:linear-killed-process} gives the killed-process last-iterate bound.
};

\node[neural, below=of nnentry] (nnproc) {
\textbf{Stopped and killed processes.}
The stopped process and Doob's inequality yield permanent stability.\\
Lemma~\ref{lem:nn_killed_process_comparison} gives the killed-process last-iterate bound.
};

\node[result, below=of linproc] (linthm) {
\textbf{Theorem~\ref{thm:log_free_permanent_stability}.}
High-probability entrance, permanent stability, and
\[
    L(W_t)
    =
    O\left(
    \frac{1}{
    \eta(t-\tau_\delta)}
    \right).
\]
};

\node[result, below=of nnproc] (nnthm) {
\textbf{Theorem~\ref{thm:NN_log_free_permanent_stability}.}
High-probability entrance, permanent stability, and
\[
    L(W_t)
    =
    O\left(
    \frac{1}{
    \eta(t-\tau_\delta^{NN})}
    \right).
\]
};

\node[
    common,
    below=6mm of
    $(linthm.south)!0.5!(nnthm.south)$,
    anchor=north
] (cor) {
\textbf{Large-stepsize acceleration.}
For $T\geq2t_{\max}$, respectively
$T\geq2t_{\max}^{NN}$, and $\eta=\Theta(T)$,
Theorems~\ref{thm:log_free_permanent_stability} and~\ref{thm:NN_log_free_permanent_stability} give
$L(W_T)=O(T^{-2})$ in
Corollaries~\ref{cor:log_free_acceleration}
and~\ref{cor:NN_log_free_acceleration}.
};

\draw[arrow]
(setup.south) -- ++(0,-2mm) -| (linprep.north);

\draw[arrow]
(setup.south) -- ++(0,-2mm) -| (nnprep.north);

\draw[arrow] (linprep) -- (linentry);
\draw[arrow] (nnprep) -- (nnentry);

\draw[arrow] (linentry) -- (linproc);
\draw[arrow] (nnentry) -- (nnproc);

\draw[arrow] (linproc) -- (linthm);
\draw[arrow] (nnproc) -- (nnthm);

\draw[arrow] (linthm.south) |- (cor.north);
\draw[arrow] (nnthm.south) |- (cor.north);

\end{tikzpicture}
\caption{
Proof dependencies for the linear and two-layer neural-network
analyses. Arrows indicate how the preparatory estimates lead to
entrance, permanent stability, last-iterate convergence, and
large-stepsize acceleration.
}
\label{fig:proof-dependencies}
\end{figure}
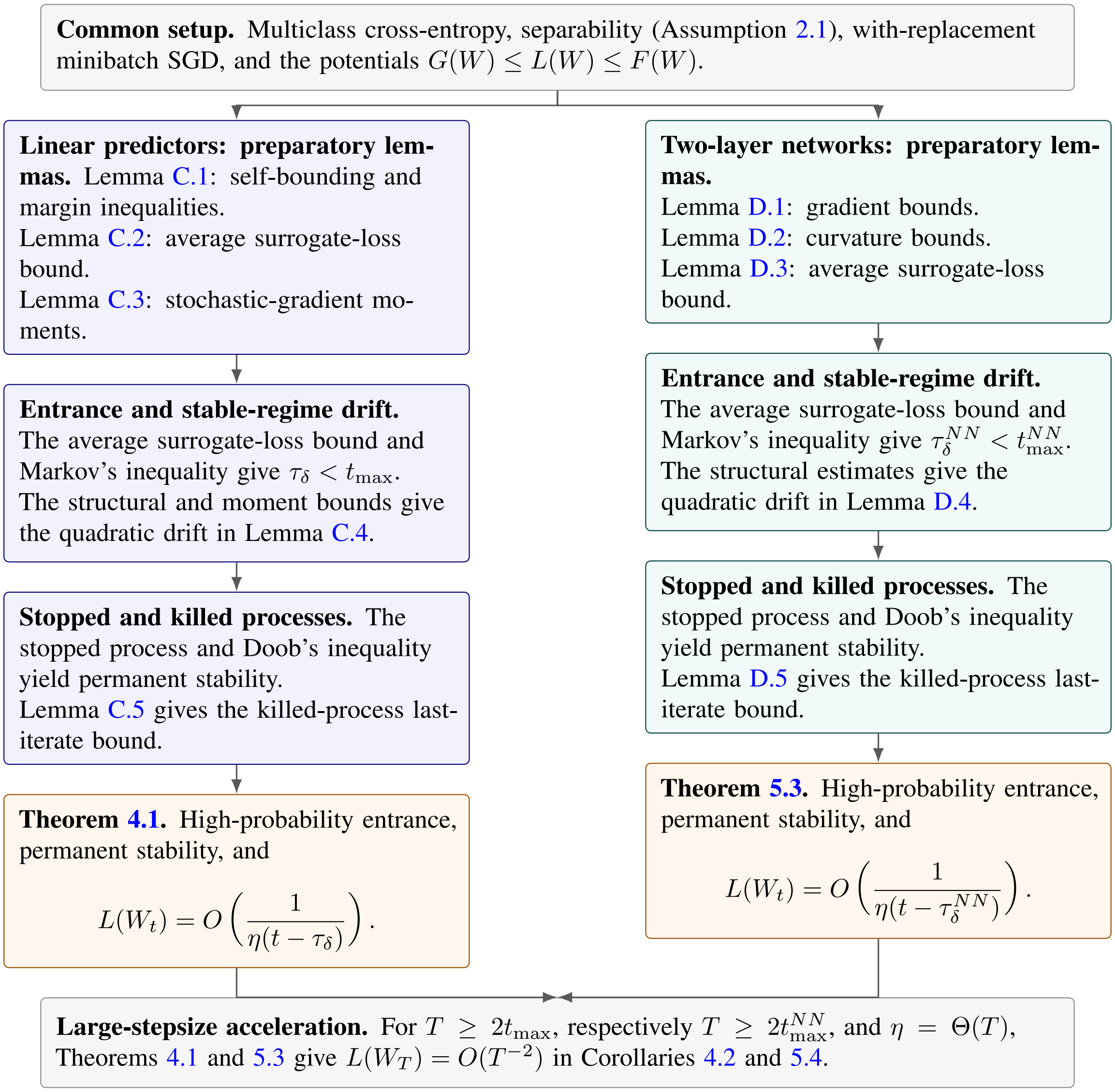
\section{Additional Related Work}
\label{app:additional-related-work}

In this appendix, we provide a more detailed discussion of related work. 
Our work is connected to several strands of literature: edge-of-stability dynamics, large-stepsize optimization for classification losses, implicit bias on separable data, and stochastic optimization beyond the small-stepsize regime.

\paragraph{Edge of stability.}
The edge-of-stability (EoS) phenomenon was systematically documented by \citet{cohen2021gradient}, who observed that neural networks trained by gradient descent often operate near, or even beyond, the classical stability threshold determined by the largest eigenvalue of the Hessian. This behavior contrasts sharply with smooth optimization theory, where stepsizes below a curvature-dependent threshold ensure monotone loss descent. In modern training, however, substantially larger stepsizes may generate short-term oscillations while preserving a decreasing trend over longer time scales.

Several works have sought to explain this apparent instability. \citet{damian2023self} identified a self-stabilization mechanism through which deterministic gradient descent can reduce its effective instability and return toward a stable region. Complementarily, \citet{cohen2024understandingoptimizationdeeplearning} introduced the central-flow perspective to describe the effective dynamics induced by large stepsizes. These analyses are primarily deterministic. For SGD, non-monotonicity may arise both from curvature-induced oscillations and from mini-batch fluctuations, making the stochastic setting qualitatively different. Our analysis accounts for this distinction through a stability criterion based on conditional Lyapunov drift rather than pathwise monotonicity.

\paragraph{Large-stepsize methods for logistic and cross-entropy losses.}
A recent line of work has shown that large stepsizes can be beneficial for classification problems with logistic-type losses. \citet{wu2024large} proved that deterministic gradient descent for logistic regression may traverse a non-monotone EoS phase before entering a stable regime with faster loss decay. \citet{cai2024large} extended this perspective to non-homogeneous two-layer networks under the binary logistic loss, establishing improved margin growth and optimization guarantees. Further refinements of large-stepsize training for logistic-type objectives were developed by \citet{wu2025large}.

Our work extends this theory along three axes. First, we study SGD rather than full-batch gradient descent and therefore separate curvature-driven instability from stochastic fluctuations. Second, we consider multiclass cross-entropy, where every example must be separated simultaneously from all competing classes and the corresponding gradients are coupled through the softmax probabilities. Third, our stable regime is defined through conditional expected descent with respect to the next mini-batch. In this sense, our results provide a stochastic and multiclass counterpart to the deterministic large-stepsize theory for logistic losses.

\paragraph{Large-stepsize methods for (strongly)-convex optimization}
Recent developments in deterministic optimization have begun to challenge the classical descent lemma viewpoint. A growing line of work \citep{pmlr-v291-bok25a,grimmer2024provably,altschuler2024acceleration,altschuler2025acceleration} has demonstrated that the stepsize does not merely regulate stability: if chosen appropriately, it can itself become a mechanism of acceleration. 
Perhaps the most striking instance of this phenomenon is given by the \emph{silver stepsize} schedule \citep{altschuler2025acceleration}, where carefully designed non-monotone stepsizes accelerate Gradient Descent (GD) on smooth convex objectives, without introducing momentum or altering the underlying iteration.

The main feature of these schemes is that some of their stepsizes lie well beyond the range suggested by the standard one-step descent argument \citep{grimmer2024provably}. Consequently, individual updates do not need to be contractive and the objective does not need to decrease monotonically along the trajectory. 
In particular, acceleration is obtained instead from the cumulative organization of a sequence of aggressive updates. More generally, these results suggest that enforcing descent at every iteration may be unnecessarily restrictive: the very instability that is excluded by the classical analysis can, when properly controlled, become an algorithmic benefit \citep{grimmer2023accelerated, zhang2026accelerated}.

\paragraph{Implicit bias in separable classification.}
Our analysis is also connected to the implicit-bias literature for separable classification. For logistic and cross-entropy losses on separable data, the empirical risk has no finite minimizer: the loss converges to zero while the parameter norm diverges. Nevertheless, the normalized parameter direction may converge to a maximum-margin solution. This phenomenon was established for linear logistic regression by \citet{soudry2018implicit} and subsequently refined and extended by \citet{ji2019implicit,nacson2019convergence}.

This geometry motivates our choice of Lyapunov function. Stability notions based on recurrence near a finite equilibrium are poorly suited to separable classification, since the parameter iterates do not converge to a finite point. The loss itself, however, converges and directly captures optimization progress. We therefore define stability through the conditional drift of the cross-entropy loss. The resulting stable regime is a loss sublevel set rather than a neighborhood of a finite minimizer, aligning the analysis with the intrinsic geometry of separable classification.

\paragraph{Stochastic optimization.}
For homogeneous linear classifiers on separable data, \citet{nacson2019stochastic} established convergence of SGD to zero loss under a fixed, non-vanishing learning rate, covering both with-replacement and without-replacement sampling. For overparameterized two-layer networks, \citet{brutzkus2018sgd} showed that SGD with sufficiently small stepsizes can learn linearly separable data. Our results complement these works by treating multiclass linear predictors and two-layer networks in the large-stepsize regime, where neither pathwise loss descent nor classical smoothness-based arguments are available.

An important direction is to extend this theory to random reshuffling and other without-replacement sampling schemes. Such schemes introduce temporal dependence across stochastic gradients and therefore require tools different from the conditional-independence arguments used for with-replacement SGD. Recent progress on random-reshuffling methods \citep{emmanouilidis2024stochastic} suggests that exploiting this dependence may lead to sharper stability and convergence guarantees for large-stepsize training.

\newpage

\section{Proof for Cross-entropy Loss}
\subsection{Notation}
For each sample $(x_i, y_i)$, let
\begin{eqnarray}
    z_i &=& W^\top x_i, \nonumber \\
    p_i &=& \mathrm{softmax}(z_i), \nonumber
\end{eqnarray}
where $p_i(j) = \frac{e^{z_i(j)}}{\sum_{\ell=1}^K e^{z_i(\ell)}}$. The gradient and Hessian of the cross-entropy loss evaluated at the sample $(x_i, y_i)$ are given by
\begin{eqnarray}
    \nabla \ell_i(W) &=& (p_i - e_{y_i})x_i^\top, \label{eq:grad-softmax}\\
    \nabla^2 \ell_i(W) &=& (\mathrm{diag}(p_i) - p_i p_i^\top) \otimes (x_i x_i^\top). \label{eq:hess-softmax}
\end{eqnarray}
Define the potential functions $F, G: \R^{d \times K} \rightarrow \R_{\geq 0}$ with
\begin{eqnarray}
F(W) &=& \frac{1}{n} \sum_{i=1}^n \sum_{j \ne y_i} e^{-\langle W({y_i}) - W(j), x_i\rangle}, \label{eq:FW}\\
G(W) &=& \frac{1}{n} \sum_{i=1}^n (1 - p_i(y_i)). \label{eq:GW}
\end{eqnarray}
Since $\log(1 + u) \le u, \forall u \in (0, 1]$, it follows that
\begin{eqnarray}
G(W) \leq L(W) \le F(W). \label{eq: L_lower_upper_bound}
\end{eqnarray}
\subsection{Preparatory Lemmas}
We first provide some preparatory lemmas that will be necessary in proving the main theorem. We start by providing a sequence of inequalities reflecting the self-bounded properties of the cross-entropy loss.

\begin{lemma} [\cite{emmanouilidis2026sgd}]\label{lemma: self_boundedness}
    For every $W \in \mathbb R^{{d} \times K}$, it holds for every $i \in [n]$ {and all indices $i_1,\ldots,i_b\in[n]$} that
    \begin{eqnarray}
    \left\|\frac{1}{b} \sum_{{r}=1}^b \nabla \ell(W^\top x_{i_r})\right\|_F
&\leq&
\frac{\sqrt{2}}{b} \sum_{{r}=1}^b\ell(W^\top x_{i_r}),
\label{eq:individual_grad_bound}
\\
\|\nabla L(W)\|_F
&\leq&
\sqrt{2}\,G(W)
\leq
\sqrt{2}\,L(W).
\label{eq:gradnorm_goal_extended}\\
\|\nabla^2 \ell(W^\top x_i)\|_{\text{op}} &\leq& 2\,\ell(W^\top x_i)\label{eq:opnorm}
\end{eqnarray}
where $G(W) = \frac{1}{n}\sum_{i=1}^n (1-p_i(y_i))$. Moreover, under Assumption~\ref{assumpt: separable} it holds that
    \begin{eqnarray}
        \|\nabla L(W)\|_F \;\ge\; \gamma\, G(W)\label{eq:grad_lower_bound}
    \end{eqnarray}
\end{lemma}

We, next, provide a bound on the potential function $G(\cdot)$ that is of critical importance for bounding quantities arising in the analysis.
\begin{lemma} [\cite{emmanouilidis2026sgd}]\label{lemma: average_G_bound}
    The iterates of \ref{SGD} with any step size $\eta > 0$ satisfy for $i \in [n]$ that
    \begin{eqnarray}
    \frac{1}{t}\sum_{k=0}^{t-1} \Expe{G(W_k)} &\leq& \frac{\sqrt{K-1}+2\ln(\gamma^2\eta t)+2\eta \left(1 + \frac{1}{b}\right)}{\eta \gamma^2 t} \label{eq:bound_G}
\end{eqnarray}
    where $G(W) = \frac{1}{n}\sum_{i=1}^n (1-p_i(y_i))$.
\end{lemma}

We, next, provide a bound from prior literature relating the norm of the stochastic gradient oracle with the norm of the full-batch gradient as well as a bound on the variance of \ref{SGD} algorithm. 

\begin{lemma}[\cite{emmanouilidis2026sgd}]\label{lemma: stochastic_grad_bound}
    The stochastic gradient oracle of the cross-entropy loss, $g_t = \frac{1}{b}\sum_{i\in\mathcal{B}_t}\nabla\ell(f(x_i;W_t),y_i)$, satisfies
    \begin{eqnarray}
        \Expep{\|g_t\|^2} &\leq& \left(1+\frac{{2}n}{b\gamma^2}\right) \|\nabla L(W_t)\|^2 \label{eq:bound_stoch_oracle}\\
        \Expe{\| g_t - \nabla L(W_t)\|^2}&\leq&\frac{2}{b}\, G(W_t),\label{eq:variance-bound_}
    \end{eqnarray}
where $G(W)
    := \frac{1}{n}
    \sum_{i=1}^n
    \Big[1 - \mathrm{softmax}(W{^\top} x_i)_{y_i}\Big]$.
\end{lemma}
We, next, establish a descent lemma for the loss inside the stable regime $\mathcal{S}$. 
\begin{lemma}[Conditional loss descent for linear predictors]
\label{lem:linear-stable-descent}
Let Assumption~\ref{assumpt: separable} hold and define
$\gamma_0:=\min\{\gamma,1\}$. Consider the iterates of
\eqref{SGD} with stepsize $\eta>0$ and batch size $b\geq 1$. Suppose that $L(W_t)\leq\widetilde L
    :=
    \min\left\{
    {\frac{1}{
    16\eta\left(
    1+\frac{2n}{b\gamma_0^2}
    \right)
    },
    \frac{1}{\sqrt{2}n\eta},
    \frac{1}{n}}
    \right\}.
    \nonumber$
Then, it holds that
\begin{eqnarray}
    \mathbb E\left[
    L(W_{t+1})-L(W_t)
    \mid\mathcal F_t
    \right]
    &\leq&
    -\frac{\eta\gamma_0^2}{8}
    L(W_t)^2.
    \label{eq:linear-stable-drift}
\end{eqnarray}
\end{lemma}

\begin{proof}
Conditionally on $\mathcal F_t$, let
$I_{t,1},\ldots,I_{t,b}$ be independent random variables
sampled uniformly from $[n]$. The minibatch gradient is
\begin{eqnarray}
    g_t
    :=
    \frac1b
    \sum_{r=1}^b
    \nabla\ell(W_t^T x_{I_{t,r}})
    \in\mathbb R^{d\times K}.
    \nonumber
\end{eqnarray}

We first control the change of the logits along the SGD update.
Fix $i\in[n]$ and define
\begin{eqnarray}
    \Delta z_i
    &:=&
    z_i(W_{t+1})-z_i(W_t)
    \nonumber\\
    &=&
    (W_{t+1}-W_t)^\top x_i
    \nonumber\\
    &\stackrel{\eqref{SGD}}{=}&
    -\eta g_t^\top x_i.
    \label{eq:linear-logit-difference-wr}
\end{eqnarray}
For every $k\in[K]$, we have
\begin{eqnarray}
    |(\Delta z_i)_k|
    &=&
    \eta
    \left|
    \left\langle
    (g_t)_k,x_i
    \right\rangle
    \right|
    \nonumber\\
    &\leq&
    \eta\|(g_t)_k\|_2\|x_i\|_2
    \nonumber\\
    &\leq&
    \eta\|g_t\|_F,
    \nonumber
\end{eqnarray}
where we used $\|x_i\|_2\leq1$. Consequently, we get
\begin{eqnarray}
    \|\Delta z_i\|_\infty
    &\leq&
    \eta\|g_t\|_F
    \nonumber\\
    &\leq&
    \frac{\eta}{b}
    \sum_{r=1}^b
    \left\|
    \nabla\ell(W_t^T x_{I_{t,r}})
    \right\|_F
    \nonumber\\
    &\leq&
    \frac{\sqrt{2}\eta}{b}
    \sum_{r=1}^b
    \ell(W_t^T x_{I_{t,r}})
    \nonumber\\
    &\leq&
    \sqrt{2}\eta
    {\max_{i\in[n]}\ell(W_t^T x_{i})}
    \nonumber\\
    &\leq&
    \sqrt{2}n\eta L(W_t),
    \label{eq:linear-logit-increment-wr}
\end{eqnarray}
where the third inequality follows from
{Lemma~\ref{lemma: self_boundedness}, inequality~\eqref{eq:individual_grad_bound},} and the last inequality follows
from ${\max_{i\in[n]}\ell(W_t^T x_{i})
    \leq
    \sum_{i=1}^n\ell(W_t^T x_{i})}
    =
    nL(W_t).$ Since $L(W_t)\leq\widetilde L$, we obtain
\begin{eqnarray}
    \|\Delta z_i\|_\infty
    &\leq&
    \sqrt{2}n\eta\widetilde L
    \nonumber\\
    &\leq&
    {1}.
    \label{eq:linear-logit-small-wr}
\end{eqnarray}

For every $\theta\in[0,1]$, define $W_t^\theta
    :=
    W_t-\theta\eta g_t.$
Since the logits are linear in $W$, we have
\begin{eqnarray}
    z_i(W_t^\theta)-z_i(W_t)
    =
    -\theta\eta g_t^\top x_i
    =
    \theta\Delta z_i.
    \nonumber
\end{eqnarray}
It follows from~\eqref{eq:linear-logit-small-wr} that
\begin{eqnarray}
    \|z_i(W_t^\theta)-z_i(W_t)\|_\infty
    \leq
    {1}.
    \nonumber
\end{eqnarray}
Thus, for every $j\neq y_i$, we obtain
\begin{eqnarray}
    z_i(W_t^\theta)_j-z_i(W_t^\theta)_{y_i}
    &\leq&
    z_i(W_t)_j-z_i(W_t)_{y_i}
    +{2}.
    \nonumber
\end{eqnarray}
Using the definition of the cross-entropy loss, we get
\begin{eqnarray}
    \ell_i(W_t^\theta)
    &=&
    \ln\left(
    1+
    \sum_{j\neq y_i}
    \exp\left(
    z_i(W_t^\theta)_j
    -
    z_i(W_t^\theta)_{y_i}
    \right)
    \right)
    \nonumber\\
    &\leq&
    \ln\left(
    1+
    {e^{2}}
    \sum_{j\neq y_i}
    \exp\left(
    z_i(W_t)_j-z_i(W_t)_{y_i}
    \right)
    \right)
    \nonumber\\
    &\leq&
    {e^{2}}\ell_i(W_t)
    \nonumber\\
    &\leq&
    {8}\ell_i(W_t).
    \label{eq:linear-loss-ratio-wr}
\end{eqnarray}
where we used $\ln(1+cu)\leq c\ln(1+u),
    c\geq1, u\geq0.$ We next apply Taylor's theorem along the SGD update. Let $\psi_i(\theta)
    :=
    \ell_i(W_t-\theta\eta g_t),
    \theta\in[0,1]$.
There exists $\theta_i\in(0,1)$ such that
\begin{eqnarray}
    \ell_i(W_{t+1})
    &=&
    \ell_i(W_t)
    -
    \eta
    \left\langle
    \nabla\ell_i(W_t),g_t
    \right\rangle
    +
    \frac{\eta^2}{2}
    \left\langle
    g_t,
    \nabla^2\ell_i(W_t^{\theta_i})g_t
    \right\rangle.
    \label{eq:linear-taylor-wr}
\end{eqnarray}
For the linear cross-entropy loss, we have from {Lemma~\ref{lemma: self_boundedness}} inequality \eqref{eq:opnorm}
\begin{eqnarray}
    \|\nabla^2\ell_i(W)\|_{\mathrm{op}}
    \leq
    2\ell_i(W).
    \nonumber
\end{eqnarray}
Therefore, using~\eqref{eq:linear-loss-ratio-wr}, we obtain
\begin{eqnarray}
    \|\nabla^2\ell_i(W_t^{\theta_i})\|_{\mathrm{op}}
    &\leq&
    2\ell_i(W_t^{\theta_i})
    \nonumber\\
    &\leq&
    {16}\ell_i(W_t).
    \label{eq:linear-hessian-shifted-wr}
\end{eqnarray}
Substituting~\eqref{eq:linear-hessian-shifted-wr} into
\eqref{eq:linear-taylor-wr}, we get
\begin{eqnarray}
    \ell_i(W_{t+1})
    &\leq&
    \ell_i(W_t)
    -
    \eta
    \left\langle
    \nabla\ell_i(W_t),g_t
    \right\rangle
    +
    {8}\eta^2\ell_i(W_t)\|g_t\|_F^2.
    \label{eq:linear-sample-descent-wr}
\end{eqnarray}
Summing~\eqref{eq:linear-sample-descent-wr} over $i\in[n]$ and
dividing by $n$, we have
\begin{eqnarray}
    L(W_{t+1})-L(W_t)
    &\leq&
    -\eta
    \left\langle
    \nabla L(W_t),g_t
    \right\rangle
    +
    {8}\eta^2L(W_t)\|g_t\|_F^2.
    \label{eq:linear-empirical-descent-wr}
\end{eqnarray}

We now take the conditional expectation. Since
$I_{t,1},\ldots,I_{t,b}$ are conditionally independent and
uniformly distributed over $[n]$, we have
$\mathbb E[g_t\mid\mathcal F_t]
    =
    \nabla L(W_t)$.
It follows from~\eqref{eq:linear-empirical-descent-wr} that
\begin{eqnarray}
    \mathbb E\left[
    L(W_{t+1})-L(W_t)
    \mid\mathcal F_t
    \right]
    &\leq&
    -\eta\|\nabla L(W_t)\|_F^2
     +
    {8}\eta^2L(W_t)
    \mathbb E\left[
    \|g_t\|_F^2
    \mid\mathcal F_t
    \right].
    \label{eq:linear-conditional-descent-wr}
\end{eqnarray}

From Lemma~\ref{lemma: stochastic_grad_bound}, we have that
\begin{eqnarray}
    \mathbb E\left[
    \|g_t\|_F^2
    \mid\mathcal F_t
    \right]
    &\leq&
    \left(
    1+\frac{2n}{b\gamma^2}
    \right)
    \|\nabla L(W_t)\|_F^2.
    \label{eq:linear-gradient-second-moment-wr}
\end{eqnarray}
Substituting~\eqref{eq:linear-gradient-second-moment-wr} into
\eqref{eq:linear-conditional-descent-wr}, we obtain
\begin{eqnarray}
    \mathbb E\left[
    L(W_{t+1})-L(W_t)
    \mid\mathcal F_t
    \right]
    &\leq&
    -\eta
    \left[
    1-
    {8}\eta
    \left(
    1+\frac{2n}{b\gamma^2}
    \right)L(W_t)
    \right]
    \|\nabla L(W_t)\|_F^2.
    \label{eq:linear-descent-final-wr}
\end{eqnarray}
Since $L(W_t)\leq\widetilde L$ and
$\gamma_0\leq\gamma$, we have
\begin{eqnarray}
    {8}\eta
    \left(
    1+\frac{2n}{b\gamma^2}
    \right)L(W_t)
    &\leq&
    {\frac{
    8\eta\left(1+\frac{2n}{b\gamma_0^2}\right)
    }{
    16\eta\left(
    1+\frac{2n}{b\gamma_0^2}
    \right)
    }
    =
    \frac12}.
    \nonumber
\end{eqnarray}
Consequently,~\eqref{eq:linear-descent-final-wr} gives
\begin{eqnarray}
    \mathbb E\left[
    L(W_{t+1})-L(W_t)
    \mid\mathcal F_t
    \right]
    &\leq&
    -\frac{\eta}{2}
    \|\nabla L(W_t)\|_F^2.
    \label{eq:linear-gradient-drift-wr}
\end{eqnarray}

It remains to lower-bound the full-gradient norm. From Lemma~\ref{lemma: self_boundedness} we have that 
\begin{eqnarray}
    \|\nabla L(W_t)\|_F
    \geq
    \gamma G(W_t).
    \nonumber
\end{eqnarray}
Since
$L(W_t)\leq\widetilde L\leq1/n$, we have that for every $i\in[n]$ it holds that
\begin{eqnarray}
    \ell_i(W_t)
    \leq
    \sum_{j=1}^n\ell_j(W_t)
    =
    nL(W_t)
    \leq1.
    \nonumber
\end{eqnarray}
Using $1-e^{-u}\geq u/2$ for $u\in[0,1]$, we obtain
\begin{eqnarray}
    G(W_t)
    &=&
    \frac1n
    \sum_{i=1}^n
    \left(
    1-e^{-\ell_i(W_t)}
    \right)
    \nonumber\\
    &\geq&
    \frac1{2n}
    \sum_{i=1}^n
    \ell_i(W_t)
    \nonumber\\
    &=&
    \frac12L(W_t)\label{eq:t_sub1}.
\end{eqnarray}
Therefore, it holds that
\begin{eqnarray}
    \|\nabla L(W_t)\|_F^2
    &\geq&
    \gamma^2G(W_t)^2
    \nonumber\\
    &\stackrel{\eqref{eq:t_sub1}}{\geq}&
    \frac{\gamma^2}{4}L(W_t)^2
    \nonumber\\
    &\geq&
    \frac{\gamma_0^2}{4}L(W_t)^2.
    \label{eq:linear-gradient-lower-bound-wr}
\end{eqnarray}
Combining~\eqref{eq:linear-gradient-drift-wr} and
\eqref{eq:linear-gradient-lower-bound-wr}, we get
\begin{eqnarray}
    \mathbb E\left[
    L(W_{t+1})-L(W_t)
    \mid\mathcal F_t
    \right]
    &\leq&
    -\frac{\eta\gamma_0^2}{8}
    L(W_t)^2.
    \nonumber
\end{eqnarray}
\end{proof}
We, next, establish the last-iterate convergence of the killed process $Z_t
    =
    L(V_t)\mathbf{1}_{\{\sigma>t\}}$, which is useful in proving the $\mathcal{O}\left(\frac{1}{t}\right)$ of \ref{SGD} inside the lower-level set in the proof of the main theorem (Theorem~\ref{thm:log_free_permanent_stability}).

\begin{lemma}[Last-iterate bound for the killed process]
\label{lem:linear-killed-process}
Let Assumption~\ref{assumpt: separable} hold.
On the event $\{\tau_\delta<\infty\}$, define
$V_t:=W_{\tau_\delta+t}$ and $\mathcal G_t:=\mathcal F_{\tau_\delta+t}$.
Let $\sigma:=\inf\left\{t\geq0:L(V_t)>\widetilde L\right\}, 
Z_t:=L(V_t)\mathbf 1_{\{\sigma>t\}}.$
Then, for every integer $T\geq1$, it holds that
\begin{eqnarray}
\mathbb E[Z_T\mid\mathcal G_0]
\leq\frac{8}{\eta\gamma_0^2T}.
\nonumber
\end{eqnarray}
\end{lemma}

\begin{proof}

Let $A_t:=\{\sigma>t\}$. Since $A_{t+1}\subseteq A_t$,
the nonnegativity of the loss and the drift inequality \eqref{eq:linear-stable-drift} from Lemma~\ref{lem:linear-stable-descent} give
\begin{eqnarray}
\mathbb E[Z_{t+1}\mid\mathcal G_t]
&=&\mathbf1_{A_t}
\mathbb E\left[L(V_{t+1})\mathbf1_{\{L(V_{t+1})\leq\widetilde L\}}
\mid\mathcal G_t\right]
\nonumber\\
&\leq&\mathbf1_{A_t}\mathbb E[L(V_{t+1})\mid\mathcal G_t]
\nonumber\\
&\leq&\mathbf1_{A_t}\left[L(V_t)-\frac{\eta\gamma_0^2}{8}L(V_t)^2\right]
\nonumber\\
&=&Z_t-\frac{\eta\gamma_0^2}{8}Z_t^2.
\nonumber
\end{eqnarray}
Define $a_t:=\mathbb E[Z_t\mid\mathcal G_0]$ and let
$C:=\eta\gamma_0^2/8$ for brevity. Taking conditional expectations and
applying Jensen's inequality we have
\begin{eqnarray}
a_{t+1}\leq a_t-Ca_t^2.
\nonumber
\end{eqnarray}
It holds that $0\leq a_t\leq a_0=L(V_0)\leq\widetilde L$ and
\begin{eqnarray}
Ca_t\leq C\widetilde L
\leq\frac{\gamma_0^2}{128\left(1+\frac{{2}n}{b\gamma_0^2}\right)}
\leq\frac{1}{128}<1.
\nonumber
\end{eqnarray}
When $a_{t+1}>0$, we obtain
\begin{eqnarray}
\frac1{a_{t+1}}
&\geq&\frac1{a_t(1-Ca_t)}
\geq\frac1{a_t}+C.
\nonumber
\end{eqnarray}
If $a_t=0$ for some $t \geq 0$, the desired bound is immediate for all later
iterations by nonnegativity and the above recursion.
Otherwise, summing the reciprocal estimate gives for all $T \geq 1$
\begin{eqnarray}
a_T
&\leq&\frac{1}{a_0^{-1}+CT}
\leq\frac1{CT}
=\frac8{\eta\gamma_0^2T}.
\nonumber
\end{eqnarray}
\end{proof}

\subsection{Proof of Theorem~\ref{thm:log_free_permanent_stability}}
\label{app:thm:entrance_permanence_stability}
\begin{yellowbox} \textbf{Proof Roadmap.} We first use the average surrogate-loss bound to show that, with
high probability, the dynamics enter the lower-level set within the
prescribed number of iterations. Starting from this region,
we apply the stable-regime drift inequality to a stopped loss process
and use Doob's maximal inequality to show that the dynamics never
leave the larger stable set. We then apply the killed-process
comparison to obtain the last-iterate convergence rate on the event
of permanent stability. Finally, we combine the entrance, stability,
and convergence events through a union bound.
\end{yellowbox}
\begin{proof}

From Lemma~\ref{lemma: average_G_bound} applied at $T = t_{\max}(\delta)$, we have that
\begin{equation}
  \frac{1}{t_{\max}}\sum_{k=0}^{t_{\max} - 1}\mathbb E[G(W_k)]
   \;\le\; \frac{\sqrt{K - 1} + 2\ln(\gamma^{2}\eta t_{\max}) + 2\eta(1 + \tfrac{1}{b})}{\eta\,\gamma^{2}\, t_{\max}}. \label{eq:lin-mc-Gbar-0}
\end{equation}
Using inequality $\ln(\gamma^{2}\eta t_{\max}) = \ln(\gamma^{2} t_{\max}) + \ln(\eta) \le \ln(\gamma^{2} t_{\max}) + \eta$,
we obtain
\begin{equation}
  \frac{1}{t_{\max}}\sum_{k=0}^{t_{\max} - 1}\mathbb E[G(W_k)]
   \;\le\; \frac{\sqrt{K - 1} + 2\ln(\gamma^{2} t_{\max}) + 2\eta(2 + \tfrac{1}{b})}{\eta\,\gamma^{2}\, t_{\max}}. \label{eq:lin-mc-Gbar-1}
\end{equation}
We, next, select $t_{\max}$ such that the right-hand side of
\eqref{eq:lin-mc-Gbar-1} 
is at most \(\delta^2\widetilde L/18\). We will show that it suffices to select
\begin{eqnarray}
    t_{\max}
    :=
    \left\lceil
    \frac1{\gamma^2}
    \max
    \left\{
        \frac{
            36\left(
                \sqrt{K-1}+2\eta\left(2+\frac1b\right)
            \right)
        }{
            \eta\delta^2\widetilde L
        },
        \frac{
            288
        }{
            \eta\delta^2\widetilde L
        }
        \ln
        \left(
            \frac{
                288
            }{
                \eta\delta^2\widetilde L
            }
        \right)
    \right\}
    \right\rceil
\label{eq:deep_entrance_time}
\end{eqnarray} We, next, verify the formula for the aforementioned $t_{\max}$:
\begin{itemize}
\item For $t \geq \dfrac{36\bigl(\sqrt{K - 1} + 2\eta(2 + \tfrac{1}{b})\bigr)}{\eta\,\delta^2\,\tilde L\,\gamma^{2}}$,
it holds that
$\dfrac{\sqrt{K - 1} + 2\eta(2 + \tfrac{1}{b})}{\eta\,\gamma^{2}\, t}
\;\le\; \dfrac{\delta^2\widetilde L}{36}$.
\item For $t \geq \dfrac{288}{\eta\,\delta^2\,\tilde L\,\gamma^{2}}\,\ln\!\Bigl(\dfrac{288}{\eta\,\delta^2\,\tilde L}\Bigr)$,
it holds that
$\gamma^{2}\, t \;\ge\; \dfrac{288}{\eta\,\delta^2\,\tilde L}\,\ln\!\Bigl(\dfrac{288}{\eta\,\delta^2\,\tilde L}\Bigr)$,
which is a sufficient condition (see Lemma~G.5 in \citet{cai2024large}) for
$\dfrac{{2}\,\ln(\gamma^{2}\, t)}{\eta\,\gamma^{2}\, t}
\;\le\; \dfrac{\delta^2\,\tilde L}{36}$.
\end{itemize} Consequently, we have
\begin{eqnarray}
    \frac1{t_{\max}}
    \sum_{k=0}^{t_{\max}-1}
    \mathbb{E}
    \left[
        G(W_k)
    \right]
    \leq
    \frac{\delta^2\widetilde L}{18}.
\label{eq:average_deep_entrance}
\end{eqnarray}
Since every term in the sum in
\eqref{eq:average_deep_entrance} is nonnegative, there exists an
iteration $t_{\mathrm{in}}
    \in
    \left\{
        0,\ldots,t_{\max}-1
    \right\}$
such that
\begin{eqnarray}
    \mathbb{E}
    \left[
        G(W_{t_{\mathrm{in}}})
    \right]
    \leq
    \frac{\delta^2\widetilde L}{18}.
\label{eq:expected_deep_entrance}
\end{eqnarray}

Let $\mathcal E_{\mathrm{in}}
    :=
    \left\{
        G(W_{t_{\mathrm{in}}})
        \leq
        \frac{\delta\widetilde L}{6}
    \right\}.$
Since \(G(W_{t_{\mathrm{in}}})\geq0\), Markov's inequality and
\eqref{eq:expected_deep_entrance} give
\begin{eqnarray}
    \mathbb{P}
    \left(
        \mathcal E_{\mathrm{in}}^c
    \right)
    &\leq&
    \frac{
        \mathbb{E}[G(W_{t_{\mathrm{in}}})]
    }{
        \delta\widetilde L/6
    }
    \leq
    \frac{
        \delta^2\widetilde L/18
    }{
        \delta\widetilde L/6
    }
    =
    \frac{\delta}{3}.
\label{eq:deep_entrance_markov}
\end{eqnarray}
Therefore, we obtain
\begin{eqnarray}
    \mathbb{P}
    \left(
        \mathcal E_{\mathrm{in}}
    \right)
    \geq
    1-\frac{\delta}{3}.
\label{eq:deep-entrance-event-probability}
\end{eqnarray}

We next show that \(\mathcal E_{\mathrm{in}}\) implies entrance into
the lower sublevel set. On \(\mathcal E_{\mathrm{in}}\), we have
\begin{eqnarray}
    G(W_{t_{\mathrm{in}}})
    \leq
    \frac{\delta\widetilde L}{6} 
    = \frac{\Tilde{L}_{low}}{2},
\label{eq:surrogate-deep-level}
\end{eqnarray}
where $\Tilde{L}_{low} = \frac{\delta\widetilde L}{3}$ is the threshold for the lower level set $\mathcal{S}_{low}$.
Since \(\Tilde{L}_{low} \leq \widetilde L\leq \frac{1}{n}\), for every \(i\in[n]\), we get
\begin{eqnarray}
    1-p_i(W_{t_{\mathrm{in}}})_{y_i}
    &\leq&
    nG(W_{t_{\mathrm{in}}})
    \nonumber\\
    &\stackrel{\eqref{eq:surrogate-deep-level}}{\leq}&
    \frac{n\Tilde{L}_{low}}{2}
    \nonumber\\
    &\leq&
    \frac12.
\label{eq:deep-entrance-classification}
\end{eqnarray}
Consequently, it holds that $p_i(W_{t_{\mathrm{in}}})_{y_i}
    \geq
    \frac12,i\in[n].$
Using $F(W)
    =
    \frac1n
    \sum_{i=1}^n
    \frac{
        1-p_i(W)_{y_i}
    }{
        p_i(W)_{y_i}
    },$
together with $p_i(W_{t_{\mathrm{in}}})_{y_i}
    \geq
    \frac12,i\in[n]$, we obtain
\begin{eqnarray}
    F(W_{t_{\mathrm{in}}})
    &\leq&
    \frac2n
    \sum_{i=1}^n
    \left(
        1-p_i(W_{t_{\mathrm{in}}})_{y_i}
    \right)
    \nonumber\\
    &=&
    2G(W_{t_{\mathrm{in}}})
    \nonumber\\
    &\stackrel{\eqref{eq:surrogate-deep-level}}{\leq}&
    \Tilde{L}_{low}.
\label{eq:F_at_deep_entrance}
\end{eqnarray}
Since \(L(W)\leq F(W)\) for every \(W \in \R^{d \cross K}\), it follows that $L(W_{t_{\mathrm{in}}})
    \leq
    \Tilde{L}_{low}.$
By the definition of \(\tau_\delta\), we thus have $\tau_\delta
    \leq
    t_{\mathrm{in}}
    <
    t_{\max}$
on \(\mathcal E_{\mathrm{in}}\). Combining this fact with
\eqref{eq:deep-entrance-event-probability}, we obtain
\begin{eqnarray}
    \mathbb{P}
    \left(
        \tau_\delta
        \leq
        t_{\max}
    \right)
    \geq
    1-\frac{\delta}{3}.
\label{eq:deep_entrance_probability}
\end{eqnarray}

We next prove that the dynamics remain permanently inside the stable
regime for \(t \geq \tau_\delta\). On the event
\(\{\tau_\delta<\infty\}\), let $V_t
    :=
    W_{\tau_\delta+t}, \mathcal G_t
    :=
    \mathcal F_{\tau_\delta+t},$
and define the exit time $\sigma
    :=
    \inf
    \{
        t\geq0:
        L(V_t)>\widetilde L
    \}.$
Since \(\tau_\delta\) is a stopping time, \(\sigma\) is a stopping time
with respect to \((\mathcal G_t)_{t\geq0}\). In particular, the events
\(\{\sigma\leq t\}\) and \(\{\sigma>t\}\) belong to
\(\mathcal G_t\).

Consider the stopped loss process $Y_t
    :=
    L(V_{t\wedge\sigma})$.
On the event \(\{\sigma\leq t\}\), the process has already been stopped,
and hence
\begin{eqnarray}
    Y_{t+1}
    =
    L(V_\sigma)
    =
    Y_t.
\label{eq:stopped-loss-after-exit}
\end{eqnarray}
On the event \(\{\sigma>t\}\), the dynamics have not left the stable
regime. Therefore, we have $L(V_t)
    \leq
    \widetilde L$
and
\begin{eqnarray}
    Y_{t+1}
    =
    L(V_{t+1}).
\end{eqnarray}
Applying the stable-regime drift inequality \eqref{eq:linear-stable-drift} on
\(\{\sigma>t\}\), we get
\begin{eqnarray}
    \mathbf 1_{\{\sigma>t\}}
    \mathbb{E}
    \left[
        L(V_{t+1})
        \mid
        \mathcal G_t
    \right]
    \leq
    \mathbf 1_{\{\sigma>t\}}L(V_t).
\label{eq:drift_before_stopping}
\end{eqnarray}
Combining \eqref{eq:stopped-loss-after-exit} and
\eqref{eq:drift_before_stopping}, we obtain
\begin{eqnarray}
    \mathbb{E}
    \left[
        Y_{t+1}
        \mid
        \mathcal G_t
    \right]
    &=&
    \mathbf 1_{\{\sigma\leq t\}}Y_t
    +
    \mathbf 1_{\{\sigma>t\}}
    \mathbb{E}
    \left[
        L(V_{t+1})
        \mid
        \mathcal G_t
    \right]
    \nonumber\\
    &\leq&
    \mathbf 1_{\{\sigma\leq t\}}Y_t
    +
    \mathbf 1_{\{\sigma>t\}}L(V_t)
    \nonumber\\
    &=&
    Y_t,
\label{eq:stopped_loss_supermartingale}
\end{eqnarray}
where the final equality follows because \(Y_t=L(V_t)\) on
\(\{\sigma>t\}\). Since \(Y_t\geq0\), we conclude that
\((Y_t,\mathcal G_t)_{t\geq0}\) is a nonnegative supermartingale.

By the definition of \(\tau_\delta\), we have
\begin{eqnarray}
    Y_0
    =
    L(W_{\tau_\delta})
    \leq
    \Tilde{L}_{low}
    =
    \frac{\delta}{3}\widetilde L.
\label{eq:stopped-loss-initial-value}
\end{eqnarray}
Moreover, if \(\sigma<\infty\), then
\begin{eqnarray}
    Y_\sigma
    =
    L(V_\sigma)
    >
    \widetilde L.
\end{eqnarray}
Thus, $\{\sigma<\infty\}
    \subseteq
    \left\{
        \sup_{t\geq0}Y_t
        \geq
        \widetilde L
    \right\}.$
Applying Doob's maximal inequality conditionally on
\(\mathcal F_{\tau_\delta}\), and then using
\eqref{eq:stopped-loss-initial-value}, we get
\begin{eqnarray}
    \mathbb{P}
    \left(
        \sigma<\infty
        \mid
        \mathcal F_{\tau_\delta}
    \right)
    &\leq&
    \mathbb{P}
    \left(
        \sup_{t\geq0}Y_t
        \geq
        \widetilde L
        \,\middle|\,
        \mathcal F_{\tau_\delta}
    \right)
    \nonumber\\
    &\stackrel{\eqref{eq:stopped-loss-initial-value}}{\leq}&
    \frac{
        Y_0
    }{
        \widetilde L
    }
    \nonumber\\
    &\leq&
    \frac{\delta}{3}.
\label{eq:deep_permanent_stability}
\end{eqnarray}
Therefore, conditionally on reaching the lower sublevel set, the
probability that the SGD dynamics ever leave the stable regime is at
most \(\delta/3\).

It remains to prove the last-iterate convergence estimate. Define the
killed process $Z_t
    :=
    L(W_{\tau_\delta+t})
    \mathbf 1_{\{\sigma>t\}}.$
By Lemma~\ref{lem:linear-killed-process}, for every admissible
integer \(T\geq1\), we have
\begin{eqnarray}
    \mathbb{E}
    \left[
        Z_T
        \mid
        \mathcal F_{\tau_\delta}
    \right]
    \leq
    \frac{8}{\eta\gamma_0^2T}.
\label{eq:killed_process_rate}
\end{eqnarray}
Since \(Z_T\geq0\), conditional Markov's inequality and
\eqref{eq:killed_process_rate} give
\begin{eqnarray}
    \mathbb{P}
    \left(
        Z_T
        >
        \frac{24}{\eta\delta\gamma_0^2T}
        \,\middle|\,
        \mathcal F_{\tau_\delta}
    \right)
    &\leq&
    \frac{
        \mathbb{E}
        \left[
            Z_T
            \mid
            \mathcal F_{\tau_\delta}
        \right]
    }{
        \frac{24}{\eta\delta\gamma_0^2T}
    }
    \nonumber\\
    &\leq&
    \frac{\delta}{3}.
\label{eq:last_iterate_markov_delta_three}
\end{eqnarray}

On the event \(\{\sigma=\infty\}\), the killed process agrees with the
original \ref{SGD} dynamics. Consequently, we have
\begin{eqnarray}
    Z_T
    =
    L(W_{\tau_\delta+T}).
\label{eq:killed-process-original-process}
\end{eqnarray}
{Let $t\geq1$ be the deterministic iteration fixed in the statement} and define $T:=t-\tau_\delta$. On the event $\mathcal E_0
    :=
    \left\{
        \tau_\delta<{t}
    \right\}$,
we have $T\geq1$. Moreover, since $\tau_\delta$ is an
$\left(\mathcal F_t\right)_{t\geq0}$-stopping time, $T$ is an
$\mathcal F_{\tau_\delta}$-measurable integer. Therefore,
Lemma~\ref{lem:linear-killed-process} can be applied conditionally with the random horizon
$T$, and we obtain
\begin{equation}
    \mathbb E\left[
        Z_{t-\tau_\delta}
        \mid
        \mathcal F_{\tau_\delta}
    \right]
    \leq
    \frac{8}{
        \eta\gamma_0^2
        \left(t-\tau_\delta\right)
    }.
\label{eq:killed-process-random-horizon}
\end{equation}
{Applying conditional Markov's inequality to \eqref{eq:killed-process-random-horizon}, as in \eqref{eq:last_iterate_markov_delta_three}, and using
\eqref{eq:killed-process-original-process}, we get}
\begin{eqnarray}
    &&
    {\mathbf 1_{\mathcal E_0}}\,
    \mathbb{P}
    \left(
        \left.
        \sigma=\infty,\,
        L(W_t)
        >
        \frac{24}{\eta\delta\gamma_0^2T}
        \,\right|\,
        \mathcal F_{\tau_\delta}
    \right)
    \leq
    \frac{\delta}{3}.
\label{eq:conditional-last-iterate-rate}
\end{eqnarray}

We finally combine the entrance, permanence, and convergence events.
Let the events
\begin{eqnarray}
    \mathcal E_{\mathrm{ent}}
    &:=&
    \left\{
        \tau_\delta
        \leq
        t_{\max}
    \right\},
\nonumber \\
    \mathcal E_{\mathrm{exit}}
    &:=&
    \left\{
        \sigma<\infty
    \right\},
\nonumber \\
    \mathcal E_{\mathrm{rate}}
    &:=&
    \left\{
        {\tau_\delta<t,}\,
        \sigma=\infty,\,
        L(W_t)
        >
        \frac{24}{\eta\delta\gamma_0^2T}
    \right\}.\nonumber
\end{eqnarray}
From \eqref{eq:deep_entrance_probability}, we have
\begin{eqnarray}
    \mathbb{P}
    \left(
        \mathcal E_{\mathrm{ent}}^c
    \right)
    \leq
    \frac{\delta}{3}.
\label{eq:entrance-failure-final}
\end{eqnarray}
Using \eqref{eq:deep_permanent_stability} and the tower property, we
obtain
\begin{eqnarray}
    \mathbb{P}
    \left(
        \mathcal E_{\mathrm{exit}}
        \cap
        \mathcal E_{\mathrm{ent}}
    \right)
    &=
    \mathbb{E}
    \left[
        \mathbf 1_{\mathcal E_{\mathrm{ent}}}
        \mathbb{P}
        \left(
            \sigma<\infty
            \mid
            \mathcal F_{\tau_\delta}
        \right)
    \right]
    \leq
    \frac{\delta}{3}.
\label{eq:exit-failure-final}
\end{eqnarray}
From \eqref{eq:conditional-last-iterate-rate} and the tower
property, we get
\begin{eqnarray}
    \mathbb{P}
    \left(
        \mathcal E_{\mathrm{rate}}
        \cap
        \mathcal E_{\mathrm{ent}}
    \right)
    &=
    \mathbb{E}
    \left[
        \mathbf 1_{\mathcal E_{\mathrm{ent}}}
        \mathbb{P}
        \left(
            \mathcal E_{\mathrm{rate}}
            \mid
            \mathcal F_{\tau_\delta}
        \right)
    \right]
    \leq
    \frac{\delta}{3}.
\label{eq:rate-failure-final}
\end{eqnarray}
Applying the union bound and using
\eqref{eq:entrance-failure-final},
\eqref{eq:exit-failure-final}, and
\eqref{eq:rate-failure-final}, we conclude that the entrance-time
bound, permanent stability, and last-iterate convergence estimate hold
simultaneously with probability at least
\begin{eqnarray}
    1
    -
    \frac{\delta}{3}
    -
    \frac{\delta}{3}
    -
    \frac{\delta}{3}
    =
    1-\delta.
\end{eqnarray}
\end{proof}

\subsection{Proof of Corollary~\ref{cor:log_free_acceleration}}
\label{app:cor:log_free_acceleration}
{The constants in Corollary~\ref{cor:log_free_acceleration} are
\[
    c_0:=\frac{\gamma_0^2\delta^2}{288\left(2+\frac1b\right)M},
    \qquad
    T_0(c):=\max\left\{4+\frac{144M}{\gamma_0^2\delta^2}\left(\sqrt{K-1}+8\ln\frac{288M}{\delta^2}\right),\frac{n}{Mc}\right\},
\]
where $M:=\max\left\{16\left(1+\frac{2n}{b\gamma_0^2}\right),\sqrt2n\right\}$.}
\begin{proof}
{We first show that $T\geq 2t_{\max}$. By the definition of $M$, the first two terms in the definition of $\widetilde L$ satisfy $\min\left\{\frac{1}{16\eta\left(1+\frac{2n}{b\gamma_0^2}\right)},\frac{1}{\sqrt2n\eta}\right\}=\frac{1}{\eta M}$, and thus $\widetilde L=\min\left\{\frac{1}{\eta M},\frac1n\right\}$. Since $T\geq T_0(c)\geq\frac{n}{Mc}$, we have $\eta=cT\geq\frac{n}{M}$, and thus $\frac{1}{\eta M}\leq\frac1{n}$. Hence, $\widetilde L=\frac{1}{\eta M}$. Using $\max\{x,y\}\leq x+y$ and $\gamma_0\leq\gamma$ in \eqref{eq:deep_entrance_time}, we get
\[
    2t_{\max}
    \leq
    2
    +
    \frac{2}{\gamma_0^2\delta^2}
    \left[
        36M\sqrt{K-1}
        +
        288M\ln\frac{288M}{\delta^2}
    \right]
    +
    \frac{144\left(2+\frac1b\right)M}{\gamma_0^2\delta^2}\,cT.
\]
Since $c\leq c_0=\frac{\gamma_0^2\delta^2}{288\left(2+\frac1b\right)M}$, the last term satisfies $\frac{144\left(2+\frac1b\right)M}{\gamma_0^2\delta^2}\,cT\leq\frac T2$. Moreover, since $T\geq T_0(c)\geq 4+\frac{144M}{\gamma_0^2\delta^2}\left(\sqrt{K-1}+8\ln\frac{288M}{\delta^2}\right)$, we have
\[
    2+\frac{2}{\gamma_0^2\delta^2}\left[36M\sqrt{K-1}+288M\ln\frac{288M}{\delta^2}\right]
    =
    \frac12\left(4+\frac{144M}{\gamma_0^2\delta^2}\left(\sqrt{K-1}+8\ln\frac{288M}{\delta^2}\right)\right)
    \leq
    \frac T2.
\]
Thus, $T\geq2t_{\max}$.}

From Theorem~\ref{thm:log_free_permanent_stability}, we have with probability at
least $1-\delta$ that $\tau_\delta
    \leq
    t_{\mathrm{max}}$.
For $T
    \geq
    2t_{\mathrm{max}},$
we obtain $\tau_\delta
    \leq
    \frac{T}{2}.$
Therefore, it holds
\begin{equation}
    T-\tau_\delta
    \geq
    \frac{T}{2}.
\label{eq:remaining_horizon_log_free}
\end{equation}

Applying Theorem~\ref{thm:log_free_permanent_stability} at iteration $T$, we
get with probability at least $1-\delta$ that
\begin{eqnarray}
    L(W_T)
    &\leq&
    \frac{
        24
    }{
        \delta{\gamma_0^2}
        \eta(T-\tau_\delta)
    }
    \nonumber\\
    &\leq&
    \frac{
        24
    }{
        \delta{\gamma_0^2}
        \eta(T/2)
    }
    \nonumber\\
    &=&
    \frac{
        48
    }{
        \delta{\gamma_0^2}\eta T
    }.
\end{eqnarray}
{Substituting $\eta=cT$}, we have with probability at least $1-\delta$ that
\[
    L(W_T)
    {\leq
    \frac{48}{\delta{\gamma_0^2}cT^2}}
    =
    \mathcal{O}\left(
        \frac1{T^2}
    \right)
\]
\end{proof}

\vspace{-0.5cm}

\section{Proofs for Two-Layer Neural Networks}\label{app:twolayer}

We provide the analysis of SGD for the two-layer
neural network~\eqref{eq: 2nn} under the multi-class cross-entropy
loss with any step size $\tilde\eta = m \eta > 0$.

\subsection{Notation}\label{app:twolayer:notation}

For each sample $(x_i, y_i)$, let
\[
  z_i(W) \;=\; \frac{1}{m}\sum_{r=1}^{m} a_r\, \phi(x_i^{\!\top} W^{(r)}) \in \mathbb{R}^{K},
  \qquad
  p_i(W) \;=\; \mathrm{softmax}\bigl(z_i(W)\bigr).
\]
The gradient and Hessian of the cross-entropy loss evaluated at the sample
$(x_i, y_i)$ are given by
\begin{align}
  \nabla \ell_i(W) &\;=\; J_i(W)^{\!\top}\bigl(p_i(W) - e_{y_i}\bigr), \label{eq:nn-grad}\\
  \nabla^{2}\ell_i(W) &\;=\; J_i(W)^{\!\top}\bigl(\mathrm{diag}(p_i(W)) - p_i(W)p_i(W)^{\!\top}\bigr) J_i(W)\nonumber \\
  &\;+\; \sum_{k=1}^{K}\bigl(p_i(W)_k - \mathbb 1_{\{k=y_i\}}\bigr)\,\nabla^{2} z_i(W)_k, \label{eq:nn-hess}
\end{align}
where $J_i(W)$ is the Jacobian of $W \mapsto z_i(W)$. We define the lifted
comparator
\[
\overline{W}_{*} \;=\; \bigl(a_1 W_*, \ldots, a_m W_*\bigr) \in \mathbb{R}^{md \times K}
\]
with $\|\overline{W}_{*}\|_F \;=\; \sqrt{m}$
and let $c_\phi := \frac{1}{m^2}\sum_{r=1}^{m} a_r^{2}=\frac{1}{m}$, since $a_r \in \{\pm 1\}$. 
We define the potential functions
$F, G : \mathbb{R}^{md \times K} \to \mathbb{R}_{\ge 0}$ as follows
\begin{align}
  F(W) &\;=\; \frac{1}{n}\sum_{i=1}^{n}\sum_{j \neq y_i}
              \exp\!\bigl(-(z_i(W)_{y_i} - z_i(W)_j)\bigr), \label{eq:nn-F}\\
  G(W) &\;=\; \frac{1}{n}\sum_{i=1}^{n}\bigl(1 - p_i(W)_{y_i}\bigr). \label{eq:nn-G}
\end{align}
Since $\log(1+u) \le u$ for $u \in (0, 1]$, it follows that
\begin{equation}
  G(W) \;\le\; L(W) \;\le\; F(W). \label{eq:nn-GLF}
\end{equation}

\subsection{Preparatory Lemmas}\label{app:twolayer:prep}

We first provide some preparatory lemmas that will be useful in the proof of the main theorem. We start by providing a lemma bounding the norm of the gradient of the loss by the loss and the potential $G(\cdot)$.

\begin{lemma}\citep{emmanouilidis2026sgd}\label{lem:nn-grad-bound}
For every $W \in \mathbb R^{md \times K}$, it holds for every $i \in [n]$ that
\begin{align}
  \|\nabla \ell_i(W)\|_F &\;\le\; \sqrt{\frac{2}{m}}\,\bigl(1 - p_i(W)_{y_i}\bigr) \;\le\; \sqrt{\frac{2}{m}}\,\ell_i(W) \;\le\; \sqrt{\frac{2}{m}}\, n\, L(W), \label{eq:nn-grad-i}\\
  \|\nabla L(W)\|_F &\;\le\; \sqrt{\frac{2}{m}}\, G(W) \;\le\; \sqrt{\frac{2}{m}}\, L(W). \label{eq:nn-grad-full}
\end{align}
Moreover, under Assumption~\ref{assumpt: separable} it holds that
\begin{equation}
  \sqrt{m}\,\|\nabla L(W)\|_F \;\ge\; \tilde \gamma\, G(W). \label{eq:nn-grad-lower}
\end{equation}
\end{lemma}

We, next, provide a Lemma establishing the self-bounded property of the cross-entropy loss with two-layer NNs.

\begin{lemma}\citep{emmanouilidis2026sgd}\label{lem:nn-hessian}
Under Assumption~\ref{assump: activation}, for any $W \in \mathbb R^{md \times K}$ and $i \in [n]$ it holds that
\begin{equation}
  \|\nabla^{2}\ell_i(W)\|_{\mathrm{op}} \;\le\; \frac{2(1+\tilde\beta)}{m}\,\ell_i(W),
  \qquad
  \|\nabla^{2} L(W)\|_{\mathrm{op}} \;\le\; \frac{2(1+\tilde\beta)}{m}\, L(W). \label{eq:nn-hessian}
\end{equation}
\end{lemma}

We next bound the potential function $G(W)$, which serves as an upper bound on critical quantities appearing in the main theorem.

\begin{lemma}\citep{emmanouilidis2026sgd}\label{lem:nn-G-cesaro}
The iterates of \ref{SGD} for the two-layer network~\eqref{eq: 2nn} with any step size
$\tilde\eta = m\eta > 0$ and batch size $b \ge 1$ satisfy
\begin{eqnarray}
    \frac{1}{t}\sum_{k=0}^{t-1}\mathbb E[G(W_k)] &\leq& \frac{2\bigl(K - 1 + 2\ln(\tilde\gamma^{2}\eta t) + 4\kappa + 2\eta(1 + \frac{1}{b})\bigr)}{\eta\tilde\gamma^2 t} \nonumber
\end{eqnarray}
where $G(W) = \frac{1}{n}\sum_{i=1}^{n}(1 - p_i(W)_{y_i})$.
\end{lemma}

We next provide a descent Lemma for the iterates inside the stable regime.
\begin{lemma}\label{lem:nn-descent}
If for some $t \ge 0$ the iterates of \ref{SGD} for the two-layer network~\eqref{eq: 2nn} satisfy
$L(W_t) \leq \tilde{L}_{NN}$,
then it holds that
\begin{equation}
  \mathbb E\left[
        L(W_{k+1})
        \mid\mathcal F_k
    \right]
    \leq
    L(W_k)
    -
    \frac{
        \eta\Tilde \gamma_0^2
    }{8}
    L(W_k)^2.\label{eq:descent_stability}
\end{equation}
\end{lemma}
\begin{proof}
    From inequality (201) in Lemma D.12 of \citet{emmanouilidis2026sgd} we have
\begin{eqnarray}
    \mathbb E\left[
        L(W_{k+1})-L(W_k)
        \mid\mathcal F_k
    \right]
    &\leq&
    -\widetilde\eta
    \left[
        1
        -
        8\eta(1+\widetilde\beta)
        \left(
            1+\frac{2n}{
                b\widetilde\gamma^2
            }
        \right)
        L(W_k)
    \right]
    \|\nabla L(W_k)\|_F^2.
    \label{eq:NN_drift_before_strict_margin}
\end{eqnarray}
If $W_k\in\mathcal S_{NN}$, then
\[
    L(W_k)
    \leq
    \frac1{
        16\eta(1+\widetilde\beta)
        \left(
            1+\frac{2n}{b\Tilde \gamma_0^2}
        \right)
    }.
\]
Since $\Tilde \gamma_0\leq\widetilde\gamma$, we have
\begin{eqnarray}
  8\eta(1+\widetilde\beta)
    \left(
        1+\frac{2n}{b\widetilde\gamma^2}
    \right)L(W_k)
    &\leq&
    8\eta(1+\widetilde\beta)
    \left(
        1+\frac{2n}{b\Tilde \gamma_0^2}
    \right)L(W_k)
    \leq
    \frac12.
\end{eqnarray}
Substituting this inequality into
\eqref{eq:NN_drift_before_strict_margin}, we obtain
\begin{equation}
    \mathbb E\left[
        L(W_{k+1})-L(W_k)
        \mid\mathcal F_k
    \right]
    \leq
    -\frac{\widetilde\eta}{2}
    \|\nabla L(W_k)\|_F^2.
    \label{eq:NN_strict_loss_drift}
\end{equation}

From Lemma~\ref{lem:nn-grad-bound}, inequality \eqref{eq:nn-grad-lower} gives
\begin{eqnarray}
    \sqrt m\|\nabla L(W_k)\|_F
    &\geq&
    \widetilde\gamma G(W_k)
    \geq
    \Tilde \gamma_0 G(W_k).
    \label{eq:NN_gradient_margin_bound}
\end{eqnarray}
Moreover, since $L(W_k)
    \leq
    \widetilde L_{NN}
    \leq
    \frac1{2ne^{\kappa+2}}
    \leq
    \frac1n$,
every individual loss satisfies
\[
    0
    \leq
    \ell_i(W_k)
    \leq
    nL(W_k)
    \leq
    1.
\]
Since $p_i(W_k)_{y_i}
    =
    e^{-\ell_i(W_k)}$
and $1-e^{-x}\geq\frac{x}{2},
    x\in[0,1],$
we obtain
\begin{eqnarray}
    G(W_k)
    &=&
    \frac1n
    \sum_{i=1}^n
    \left(
        1-e^{-\ell_i(W_k)}
    \right)
    \nonumber\\
    &\geq&
    \frac1{2n}
    \sum_{i=1}^n
    \ell_i(W_k)
    =
    \frac12L(W_k).
    \label{eq:NN_G_lower_bound}
\end{eqnarray}
Combining \eqref{eq:NN_gradient_margin_bound} and
\eqref{eq:NN_G_lower_bound}, we obtain
\[
    \sqrt m\|\nabla L(W_k)\|_F
    \geq
    \frac{\Tilde \gamma_0}{2}L(W_k),
\]
and therefore
\begin{equation}
    \|\nabla L(W_k)\|_F^2
    \geq
    \frac{
        \Tilde \gamma_0^2
    }{4m}
    L(W_k)^2.
    \label{eq:NN_gradient_lower_bound_loss}
\end{equation}
Substituting \eqref{eq:NN_gradient_lower_bound_loss} into
\eqref{eq:NN_strict_loss_drift}, and using
$\widetilde\eta=m\eta$, we get
\begin{equation}
    \mathbb E\left[
        L(W_{k+1})
        \mid\mathcal F_k
    \right]
    \leq
    L(W_k)
    -
    \frac{
        \eta\Tilde \gamma_0^2
    }{8}
    L(W_k)^2.
    \label{eq:NN_quadratic_loss_drift}
\end{equation}
\end{proof}
\begin{lemma}[Last-iterate bound for the killed process]
\label{lem:nn_killed_process_comparison}
Condition on the event $\{{\tau_\delta^{NN}}<\infty\}$, and define
$V_t:=W_{{\tau_\delta^{NN}}+t}, \mathcal{G}_t:=\mathcal{F}_{{\tau_\delta^{NN}}+t}$.
Let
$\sigma
    :=
    \inf\left\{
        {t\geq 0}:
        L(V_t)>{\widetilde L_{NN}}
    \right\}$
be the first exit time from the stable set after ${\tau_\delta^{NN}}$, and
define the killed loss process
$Z_t
    :=
    L(V_t)\mathbf{1}_{\{\sigma>t\}}.$
Then, {for every integer $t\geq1$, on the event $\{\tau_\delta^{NN}<t\}$,} we have
\begin{equation}
    \mathbb E\left[
        Z_{t-\tau_\delta^{NN}}
        \mid
        \mathcal F_{\tau_\delta^{NN}}
    \right]
    \leq
    \frac8{
        \eta\Tilde \gamma_0^2
        (t-\tau_\delta^{NN})
    }.
\label{eq:killed_process_last_iterate_NN}
\end{equation}
\end{lemma}

\begin{proof}
    Let $A_t:=\{\sigma>t\}, Z_t:=L(V_t)\mathbf 1_{A_t}$.
Since $A_{t+1}
    =
    A_t
    \cap
    \left\{
        L(V_{t+1})
        \leq
        \widetilde L_{NN}
    \right\},$
the nonnegativity of the loss and
\eqref{eq:descent_stability} imply
\begin{eqnarray}
    \mathbb E\left[
        Z_{t+1}\mid\mathcal G_t
    \right]
    &\leq&
    \mathbf 1_{A_t}
    \mathbb E\left[
        L(V_{t+1})
        \mid\mathcal G_t
    \right]
    \nonumber\\
    &\leq&
    \mathbf 1_{A_t}
    \left[
        L(V_t)
        -
        \frac{
            \eta\Tilde \gamma_0^2
        }{8}
        L(V_t)^2
    \right]
    \nonumber\\
&=&
    Z_t
    -
    \frac{
        \eta\Tilde \gamma_0^2
    }{8}
    Z_t^2.
    \label{eq:NN_killed_quadratic_recursion}
\end{eqnarray}

Define $a_t
    :=
    \mathbb E\left[
        Z_t\mid\mathcal G_0
    \right],$
$C_{NN}
    :=
    \frac{
        \eta\Tilde \gamma_0^2
    }{8}.$
Taking conditional expectation in
\eqref{eq:NN_killed_quadratic_recursion} and using conditional
Jensen's inequality, we obtain
\begin{equation}
    a_{t+1}
    \leq
    a_t-C_{NN}a_t^2.
    \label{eq:NN_expected_quadratic_recursion}
\end{equation}
We have that $a_t
    \leq
    a_0
    \leq
    \widetilde L_{NN}.$
Moreover, it holds
\begin{eqnarray}
    C_{NN}a_t
    &\leq&
    C_{NN}
    \widetilde L_{NN}
    \nonumber\\
    &\leq&
    \frac{
        \eta\Tilde \gamma_0^2
    }{8}
    \frac1{
        16\eta(1+\widetilde\beta)
        \left(
            1+\frac{2n}{b\Tilde \gamma_0^2}
        \right)
    }
    \nonumber\\
    &\leq&
    \frac1{128}
    <
    1.
\end{eqnarray}
If $a_{t+1}>0$, then
\eqref{eq:NN_expected_quadratic_recursion} gives
\begin{eqnarray}
    \frac1{a_{t+1}}
    &\geq&
    \frac1{
        a_t(1-C_{NN}a_t)
    }
    \nonumber\\
    &\geq&
    \frac{
        1+C_{NN}a_t
    }{a_t}
    =
    \frac1{a_t}
    +
    C_{NN},
\end{eqnarray}
where we used $(1-x)^{-1}\geq1+x$ for $x\in[0,1)$. If
$a_{t+1}=0$, the desired upper bound is immediate. Unrolling the
aforementioned inequality gives, for every integer $T\geq1$, that
\begin{equation}
    a_T
    \leq
    \frac1{
        a_0^{-1}+C_{NN}T
    }
    \leq
    \frac1{
        C_{NN}T
    }
    =
    \frac8{
        \eta\Tilde \gamma_0^2T
    }.
    \label{eq:NN_expected_log_free_rate}
\end{equation}

{Fix an integer
$t\geq1$. On the event $\left\{
        \tau_\delta^{NN}
        <
        t
    \right\},$}
we have $t-\tau_\delta^{NN}\geq1$. Applying
\eqref{eq:NN_expected_log_free_rate} conditionally for each
possible value of $\tau_\delta^{NN}$, we obtain
\begin{equation}
    \mathbb E\left[
        Z_{t-\tau_\delta^{NN}}
        \mid
        \mathcal F_{\tau_\delta^{NN}}
    \right]
    \leq
    \frac8{
        \eta\Tilde \gamma_0^2
        (t-\tau_\delta^{NN})
    }.
\end{equation}
\end{proof}

\subsection{Proof for Theorem~\ref{thm:NN_log_free_permanent_stability}}
\label{app: thm:NN_log_free_permanent_stability}
\begin{yellowbox} \textbf{Proof Roadmap.} We first use the average surrogate-loss estimate for the neural
network to prove high-probability entrance into the lower-level set.
After shifting the process to the entrance time, we combine the
stable-regime loss drift with a stopped-process argument to show that
the dynamics remain permanently inside $\mathcal S_{NN}$. We then
apply the killed-process comparison to obtain the last-iterate rate
on the event of no exit. Finally, a union bound combines the entrance,
permanent-stability, and convergence guarantees.
\end{yellowbox}
\begin{proof}
\textbf{Entrance into the lower sublevel set $\mathcal{S}_{low}$.}
From Lemma~\ref{lem:nn-G-cesaro} {and the inequality $\ln(\widetilde\gamma^2\eta T)\leq\ln(\widetilde\gamma^2T)+\eta$}, for every integer $T\geq1$, we have
\begin{equation}
    \frac1T
    \sum_{k=0}^{T-1}
    \mathbb E\left[G(W_k)\right]
    \leq
    \frac{
        2\left(
            K-1
            +
            2\ln(\widetilde\gamma^2T)
            +
            4\kappa
            +
            2\eta\left(2+\frac1b\right)
        \right)
    }{
        \eta\widetilde\gamma^2T
    }.
    \label{eq:NN_existing_average_G}
\end{equation}
For $\delta\in(0,1)$, define $t_{\max}^{NN}
    :=
    \left\lceil
    \max\left\{
    \frac{
        72\left[
            K-1
            +
            4\kappa
            +
            2\eta\left(2+\frac1b\right)
        \right]
    }{
        \eta\delta^2
        \widetilde L_{NN}
        \widetilde\gamma^2
    },
    \frac{288}{
        \eta\delta^2
        \widetilde L_{NN}
        \widetilde\gamma^2
    }
    \ln\left(
        \frac{288}{
            \eta\delta^2
            \widetilde L_{NN}
        }
    \right)
    \right\}
    \right\rceil .$
We verify that this choice makes the right-hand side of
\eqref{eq:NN_existing_average_G} at most
$\delta^2\widetilde L_{NN}/18$.

\begin{itemize}
    \item For
    $T\geq
        \frac{
            72\left[
                K-1
                +
                4\kappa
                +
                2\eta\left(2+\frac1b\right)
            \right]
        }{
            \eta\delta^2
            \widetilde L_{NN}
            \widetilde\gamma^2
        },$
    we have $\frac{
            2\left[
                K-1
                +
                4\kappa
                +
                2\eta\left(2+\frac1b\right)
            \right]
        }{
            \eta\widetilde\gamma^2T
        }
        \leq
        \frac{
            \delta^2\widetilde L_{NN}
        }{36}.$

    \item For $  T
        \geq
        \frac{288}{
            \eta\delta^2
            \widetilde L_{NN}
            \widetilde\gamma^2
        }
        \ln\left(
            \frac{288}{
                \eta\delta^2
                \widetilde L_{NN}
            }
        \right),$
    define $
        Q_\delta^{NN}
        :=
        \frac{288}{
            \eta\delta^2
            \widetilde L_{NN}
        }$.
    Then $\widetilde\gamma^2T
        \geq
        Q_\delta^{NN}
        \ln\left(Q_\delta^{NN}\right).$
    Moreover, the definition of $\widetilde L_{NN}$ gives
    $\eta\widetilde L_{NN}\leq1/16$, and hence
    $Q_\delta^{NN}\geq4608$. Thus, $Q_\delta^{NN}
        \ln\left(Q_\delta^{NN}\right)>e$.
    Since $x\mapsto\ln(x)/x$ is decreasing for $x\geq e$, we obtain
    \begin{eqnarray}
        \frac{
            4\ln(\widetilde\gamma^2T)
        }{
            \eta\widetilde\gamma^2T
        }
        &\leq&\qquad
        \frac{
            4\ln\left(
                Q_\delta^{NN}
                \ln(Q_\delta^{NN})
            \right)
        }{
            \eta Q_\delta^{NN}
            \ln(Q_\delta^{NN})
        }
        \nonumber\\
    &\leq&\qquad
        \frac{
            8\ln(Q_\delta^{NN})
        }{
            \eta Q_\delta^{NN}
            \ln(Q_\delta^{NN})
        }
        \nonumber\\
    &=&\qquad
        \frac8{
            \eta Q_\delta^{NN}
        }
        =
        \frac{
            \delta^2\widetilde L_{NN}
        }{36}.
        \label{eq:NN_logarithmic_entrance}
    \end{eqnarray}
\end{itemize}

Applying \eqref{eq:NN_existing_average_G} with
$T=t_{\max}^{NN}$, and using
the aforementioned bounds, we obtain
\begin{eqnarray}
    \frac1{t_{\max}^{NN}}
    \sum_{k=0}^{t_{\max}^{NN}-1}
    \mathbb E\left[G(W_k)\right]
    &\leq&
    \frac{
        \delta^2\widetilde L_{NN}
    }{36}
    +
    \frac{
        \delta^2\widetilde L_{NN}
    }{36}
    \nonumber\\
    &=&
    \frac{
        \delta^2\widetilde L_{NN}
    }{18}.
    \label{eq:NN_average_G_deep_entrance}
\end{eqnarray}
Thus, there exists an iteration
$t_{\mathrm{in}}^{NN}
    \in
    \left\{
        0,\ldots,t_{\max}^{NN}-1
    \right\}$
such that
\begin{equation}
    \mathbb E\left[
        G(W_{t_{\mathrm{in}}^{NN}})
    \right]
    \leq
    \frac{
        \delta^2\widetilde L_{NN}
    }{18}.
    \label{eq:NN_expected_G_deep_entrance}
\end{equation}
Let $\mathcal E_{\mathrm{in}}^{NN}
    :=
    \left\{
        G(W_{t_{\mathrm{in}}^{NN}})
        \leq
        \frac{
            \delta\widetilde L_{NN}
        }{6}
    \right\}.$
Since $G(W_{t_{\mathrm{in}}^{NN}})\geq0$, Markov's
inequality gives
\begin{eqnarray}
    \mathbb P\left(
        \left(
            \mathcal E_{\mathrm{in}}^{NN}
        \right)^c
    \right)
    &\leq&
    \frac{
        \mathbb E[
            G(W_{t_{\mathrm{in}}^{NN}})
        ]
    }{
        \delta\widetilde L_{NN}/6
    }
    \nonumber\\
    &\leq&
    \frac{
        \delta^2\widetilde L_{NN}/18
    }{
        \delta\widetilde L_{NN}/6
    }
    =
    \frac{\delta}{3}.
    \label{eq:NN_deep_entrance_markov}
\end{eqnarray}

On $\mathcal E_{\mathrm{in}}^{NN}$, for every $i\in[n]$,
we have
\begin{eqnarray}
    1-p_i(W_{t_{\mathrm{in}}^{NN}})_{y_i}
    &\leq&
    nG(W_{t_{\mathrm{in}}^{NN}})
    \nonumber\\
    &\leq&
    \frac{
        n\delta\widetilde L_{NN}
    }{6}
    \nonumber\\
    &\leq&
    \frac{\delta}{
        12e^{\kappa+2}
    }
    <
    \frac12,
\end{eqnarray}
where we used the fact that $\widetilde L_{NN}
    \leq
    \frac1{2ne^{\kappa+2}}$ at the last step.
Hence, it holds $p_i(W_{t_{\mathrm{in}}^{NN}})_{y_i}
    \geq
    \frac12.$
Using
$L(W)
    \leq
    F(W)
    =
    \frac1n
    \sum_{i=1}^n
    \frac{
        1-p_i(W)_{y_i}
    }{
        p_i(W)_{y_i}
    },$
we obtain
\begin{eqnarray}
    L(W_{t_{\mathrm{in}}^{NN}})
    &\leq&
    F(W_{t_{\mathrm{in}}^{NN}})
    \nonumber\\
    &\leq&
    \frac2n
    \sum_{i=1}^n
    \left(
        1-p_i(W_{t_{\mathrm{in}}^{NN}})_{y_i}
    \right)
    \nonumber\\
    &=&
    2G(W_{t_{\mathrm{in}}^{NN}})
    \leq
    \frac{
        \delta\widetilde L_{NN}
    }{3}.
    \label{eq:NN_deep_entrance_loss}
\end{eqnarray}

Define $\tau_\delta^{NN}
    :=
    \inf\left\{
        k\geq0:
        L(W_k)
        \leq
        \frac{
            \delta\widetilde L_{NN}
        }{3}
    \right\}$. On $\mathcal E_{\mathrm{in}}^{NN}$, we have $\tau_\delta^{NN}
    \leq
    t_{\mathrm{in}}^{NN}
    \leq
    t_{\max}^{NN}-1.$
Combining this inequality with
\eqref{eq:NN_deep_entrance_markov}, we conclude that
\begin{equation}
    \mathbb P\left(
        \tau_\delta^{NN}
        <
        t_{\max}^{NN}
    \right)
    \geq
    1-\frac{\delta}{3}.
    \label{eq:NN_deep_entrance_probability}
\end{equation}

\paragraph{Loss drift inside the stable set.}
Let that $\Tilde \gamma_0
    :=
    \min\left\{
        \Tilde\gamma,1
    \right\}$.
From Lemma~\ref{lem:nn-descent}, on the event
$\{L(W_k)\leq\widetilde L_{NN}\}$,we get that
\begin{equation}
    \mathbb E\left[
        L(W_{k+1})
        \mid\mathcal F_k
    \right]
    \leq
    L(W_k)
    -
    \frac{
        \eta\Tilde \gamma_0^2
    }{8}
    L(W_k)^2.
    \label{eq:NN_quadratic_loss_drift}
\end{equation}

\paragraph{Permanent stability after entrance.}
Condition on $\{\tau_\delta^{NN}<\infty\}$ and define
$V_t
    :=
    W_{\tau_\delta^{NN}+t},
    \mathcal G_t
    :=
    \mathcal F_{\tau_\delta^{NN}+t}.$
Define $\sigma
    :=
    \inf\left\{
        t\geq0:
        L(V_t)>
        \widetilde L_{NN}
    \right\}$
and $Y_t
    :=
    L(V_{t\wedge\sigma}).$
Since $\sigma$ is a stopping time with respect to
$(\mathcal G_t)_{t\geq0}$, the events
$\{\sigma\leq t\}$ and $\{\sigma>t\}$ belong to $\mathcal G_t$.

On $\{\sigma\leq t\}$, the process has already stopped, and hence we have
\[
    Y_{t+1}
    =
    L(V_\sigma)
    =
    Y_t.
\]
On $\{\sigma>t\}$, the dynamics have not exited the stable set, so it holds that $L(V_t)
    \leq
    \widetilde L_{NN}$ and
\[
 Y_{t+1}=L(V_{t+1}).
\]
Therefore, we have
\begin{eqnarray}
    \mathbb E\left[
        Y_{t+1}
        \mid\mathcal G_t
    \right]
    &=&
    \mathbf 1_{\{\sigma\leq t\}}Y_t
    +
    \mathbf 1_{\{\sigma>t\}}
    \mathbb E\left[
        L(V_{t+1})
        \mid\mathcal G_t
    \right]
    \nonumber\\
    &\leq&
    \mathbf 1_{\{\sigma\leq t\}}Y_t
    +
    \mathbf 1_{\{\sigma>t\}}L(V_t)
    \nonumber\\
    &=&
    Y_t,
    \label{eq:NN_stopped_supermartingale}
\end{eqnarray}
where the inequality follows from
\eqref{eq:NN_quadratic_loss_drift} on $\{\sigma>t\}$. Thus,
$(Y_t,\mathcal G_t)_{t\geq0}$ is a nonnegative supermartingale.

It holds that $Y_0
    =
    L(W_{\tau_\delta^{NN}})
    \leq
    \frac{
        \delta\widetilde L_{NN}
    }{3}$.
For every integer $T\geq1$, Doob's maximal inequality gives
\begin{eqnarray}
    \mathbb P\left(
        \sigma\leq T
        \mid
        \mathcal F_{\tau_\delta^{NN}}
    \right)
    &\leq&
    \mathbb P\left(
        \max_{0\leq s\leq T}Y_s
        \geq
        \widetilde L_{NN}
        \;\middle|\;
        \mathcal F_{\tau_\delta^{NN}}
    \right)
    \nonumber\\
    &\leq&
    \frac{
        L(W_{\tau_\delta^{NN}})
    }{
        \widetilde L_{NN}
    }
    \leq
    \frac{\delta}{3}.
\end{eqnarray}
Letting $T\to\infty$, we obtain
\begin{equation}
    \mathbb P\left(
        \sigma<\infty
        \mid
        \mathcal F_{\tau_\delta^{NN}}
    \right)
    \leq
    \frac{\delta}{3}.
    \label{eq:NN_permanent_stability_probability}
\end{equation}

\paragraph{Last-iterate convergence.}
{Let $t\geq1$ be the integer fixed in the statement. On
$\{\tau_\delta^{NN}<t\}$, we have}
$t-\tau_\delta^{NN}\geq1$.
From Lemma~\ref{lem:nn_killed_process_comparison}, we have,
on $\{\tau_\delta^{NN}<t\}$,
\begin{eqnarray}
    \mathbb E\left[
        Z_{t-\tau_\delta^{NN}}
        \mid
        \mathcal F_{\tau_\delta^{NN}}
    \right]
    \leq
    \frac8{
        \eta\widetilde\gamma_0^2
        (t-\tau_\delta^{NN})
    }.
    \nonumber
\end{eqnarray}
Conditional Markov's inequality thus gives
\begin{equation}
    \mathbb P\left(
        Z_{t-\tau_\delta^{NN}}
        >
        \frac{24}{
            \delta\eta\Tilde \gamma_0^2
            (t-\tau_\delta^{NN})
        }
        \;\middle|\;
        \mathcal F_{\tau_\delta^{NN}}
    \right)
    \leq
    \frac{\delta}{3}.
    \label{eq:NN_log_free_markov}
\end{equation}
On $\{\sigma=\infty\}$, the killed process agrees with the actual
SGD dynamics, so that $Z_{t-\tau_\delta^{NN}}
    =
    L(W_t).$
Thus, it holds that
\begin{equation}
    {\mathbf 1_{\{\tau_\delta^{NN}<t\}}}\,
    \mathbb P\left(
        \sigma=\infty,\,
        L(W_t)>
        \frac{24}{
            \delta\eta\Tilde \gamma_0^2
            (t-\tau_\delta^{NN})
        }
        \;\middle|\;
        \mathcal F_{\tau_\delta^{NN}}
    \right)
    \leq
    \frac{\delta}{3}.
    \label{eq:NN_actual_rate_failure}
\end{equation}

\paragraph{Union bound.}
{Let $\mathcal E_0^{NN}:=\{\tau_\delta^{NN}<t_{\max}^{NN}\}$.}
From \eqref{eq:NN_deep_entrance_probability}, it holds that
\[
    \mathbb P\left(
        \left(
            \mathcal E_0^{NN}
        \right)^c
    \right)
    \leq
    \frac{\delta}{3}.
\]
Moreover, by \eqref{eq:NN_permanent_stability_probability} and the
tower property, we get
\[
    \mathbb P\left(
        \mathcal E_0^{NN}
        \cap
        \{\sigma<\infty\}
    \right)
    \leq
    \frac{\delta}{3}.
\]
From \eqref{eq:NN_actual_rate_failure} and the tower
property, we have
\begin{eqnarray}
&&
    \mathbb P\left(
        \mathcal E_0^{NN},\,
        {\tau_\delta^{NN}<t,}\,
        \sigma=\infty,\,
        L(W_t)>
        \frac{24}{
            \delta\eta\Tilde \gamma_0^2
            (t-\tau_\delta^{NN})
        }
    \right)
    \leq
    \frac{\delta}{3}.
\end{eqnarray}
A union bound over these three failure events shows that, with
probability at least $1-\delta$, it holds $\tau_\delta^{NN}
    <
    t_{\max}^{NN}$ and $L(W_k)
    \leq
    \widetilde L_{NN}$ for every $k\geq\tau_\delta^{NN}$
and, {if $\tau_\delta^{NN}<t$}, it holds that
\[
    L(W_t)
    \leq
    \frac{24}{
        \delta\eta\Tilde \gamma_0^2
        (t-\tau_\delta^{NN})
    }.
\]
\end{proof}

\subsection{Proof of Corollary~\ref{cor:NN_log_free_acceleration}}
\label{app:cor:NN_log_free_acceleration}
{The constants in Corollary~\ref{cor:NN_log_free_acceleration} are $c_0^{NN}:=\frac{\tilde\gamma_0^2\delta^2}{576\left(2+\frac1b\right)M_{NN}}$ and
\[
    T_0^{NN}(c):=\max\left\{4+\frac{288M_{NN}}{\tilde\gamma_0^2\delta^2}\left(K-1+4\kappa+4\ln\frac{288M_{NN}}{\delta^2}\right),\frac{2ne^{\kappa+2}}{M_{NN}c}\right\},
\]
where $M_{NN}:=\max\left\{16(1+\widetilde\beta)\left(1+\frac{2n}{b\tilde\gamma_0^2}\right),\sqrt2n\right\}$.}

\begin{proof}
{We first show that $T\geq 2t_{\max}^{NN}$. By the definition of $M_{NN}$ and since $\tilde\gamma_0^2=\min\{\tilde\gamma^2,1\}$, the first two terms in \eqref{eq:NN_quantitative_stable_threshold} satisfy $\min\left\{\frac{1}{16\eta(1+\widetilde\beta)\left(1+\frac{2n}{b\tilde\gamma_0^2}\right)},\frac{1}{\sqrt2n\eta}\right\}=\frac{1}{\eta M_{NN}}$, and thus $\widetilde L_{NN}=\min\left\{\frac{1}{\eta M_{NN}},\frac{1}{2ne^{\kappa+2}}\right\}$. Since $T\geq T_0^{NN}(c)\geq\frac{2ne^{\kappa+2}}{M_{NN}c}$, we have $\eta=cT\geq\frac{2ne^{\kappa+2}}{M_{NN}}$, and thus $\frac{1}{\eta M_{NN}}\leq\frac{1}{2ne^{\kappa+2}}$. Hence, $\widetilde L_{NN}=\frac{1}{\eta M_{NN}}$. Using $\max\{x,y\}\leq x+y$ and $\tilde\gamma_0\leq\tilde\gamma$ in the definition of $t_{\max}^{NN}$, we get
\[
    2t_{\max}^{NN}
    \leq
    2
    +
    \frac{2}{\tilde\gamma_0^2\delta^2}
    \left[
        72M_{NN}(K-1+4\kappa)
        +
        288M_{NN}\ln\frac{288M_{NN}}{\delta^2}
    \right]
    +
    \frac{288\left(2+\frac1b\right)M_{NN}}{\tilde\gamma_0^2\delta^2}\,cT.
\]
Since $c\leq c_0^{NN}=\frac{\tilde\gamma_0^2\delta^2}{576\left(2+\frac1b\right)M_{NN}}$, the last term satisfies $\frac{288\left(2+\frac1b\right)M_{NN}}{\tilde\gamma_0^2\delta^2}\,cT\leq\frac T2$. Moreover, since $T\geq T_0^{NN}(c)\geq 4+\frac{288M_{NN}}{\tilde\gamma_0^2\delta^2}\left(K-1+4\kappa+4\ln\frac{288M_{NN}}{\delta^2}\right)$, we have
\begin{align*}
    &2+\frac{2}{\tilde\gamma_0^2\delta^2}\left[72M_{NN}(K-1+4\kappa)+288M_{NN}\ln\frac{288M_{NN}}{\delta^2}\right]\\
    &\qquad=
    \frac12\left(4+\frac{288M_{NN}}{\tilde\gamma_0^2\delta^2}\left(K-1+4\kappa+4\ln\frac{288M_{NN}}{\delta^2}\right)\right)
    \leq
    \frac T2.
\end{align*}
Thus, $T\geq2t_{\max}^{NN}$.}

From Theorem~\ref{thm:NN_log_free_permanent_stability}, we get that with probability at least
$1-\delta$, it holds $\tau_\delta^{\mathrm{NN}}
    <
    t_{\max}^{\mathrm{NN}}$
and
\begin{equation}
    L(W_T)
    \leq
    \frac{24}{
        \delta\eta\Tilde \gamma_0^2
        \left(
            T-\tau_\delta^{\mathrm{NN}}
        \right)
    }.
    \label{eq:NN_acceleration_initial_bound}
\end{equation}
Since $T
    \geq
    2t_{\max}^{\mathrm{NN}}$
and
$\tau_\delta^{\mathrm{NN}}
    <
    t_{\max}^{\mathrm{NN}}$
we obtain
\begin{eqnarray}
    T-\tau_\delta^{\mathrm{NN}}
    &>&
    T-t_{\max}^{\mathrm{NN}}(\delta,\eta)
    \nonumber\\
    &\geq&
    T-\frac{T}{2}
    =
    \frac{T}{2}.
\end{eqnarray}
Substituting this estimate into
\eqref{eq:NN_acceleration_initial_bound}, we get
\begin{eqnarray}
    L(W_T)
    &\leq&
    \frac{48}{
        \delta\eta\Tilde \gamma_0^2T
    }.
\end{eqnarray}
Finally, {substituting $\eta=cT$}, we conclude with probability at least
$1-\delta$, it holds 
\[
    L(W_T)
    {\leq
    \frac{48}{\delta\tilde\gamma_0^2cT^2}}
    =
    \mathcal O\left(
        \frac1{T^2}
    \right).
\]
\end{proof}

\end{document}